\documentclass{article} 
\usepackage[preprint]{colm2026_conference}

\usepackage{microtype}
\usepackage{hyperref}
\usepackage{url}
\usepackage{booktabs}
\usepackage{graphicx}
\usepackage{longtable}
\usepackage{placeins}
\usepackage{mdframed}
\usepackage{fvextra}
\usepackage{amssymb}
\usepackage{newunicodechar}
\newunicodechar{✓}{\checkmark}
\usepackage{listings}
\usepackage{tcolorbox}
\usepackage{amssymb}
\usepackage{amsmath}

\mdfdefinestyle{promptblock}{
  backgroundcolor=black!4,
  linecolor=black!35,
  linewidth=0.6pt,
  roundcorner=2pt,
  skipabove=6pt,
  skipbelow=6pt,
  innerleftmargin=6pt,
  innerrightmargin=6pt,
  innertopmargin=6pt,
  innerbottommargin=6pt
}
\DefineVerbatimEnvironment{verbatim}{Verbatim}{
  breaklines=true,
  breaksymbolleft={},
  breaksymbolright={},
  breakautoindent=false,
  xleftmargin=2pt,
  xrightmargin=2pt,
  fontsize=\scriptsize
}
\surroundwithmdframed[style=promptblock]{verbatim}

\mdfdefinestyle{transcriptblock}{
  backgroundcolor=blue!3,
  linecolor=blue!45!black,
  linewidth=0.7pt,
  roundcorner=2pt,
  skipabove=6pt,
  skipbelow=6pt,
  innerleftmargin=6pt,
  innerrightmargin=6pt,
  innertopmargin=6pt,
  innerbottommargin=6pt
}
\DefineVerbatimEnvironment{transcript}{Verbatim}{
  breaklines=true,
  breaksymbolleft={},
  breaksymbolright={},
  breakautoindent=false,
  commandchars=\\\{\},
  xleftmargin=2pt,
  xrightmargin=2pt,
  fontsize=\scriptsize
}
\surroundwithmdframed[style=transcriptblock]{transcript}

\usepackage{lineno}

\definecolor{darkblue}{rgb}{0, 0, 0.5}
\hypersetup{colorlinks=true, citecolor=darkblue, linkcolor=darkblue, urlcolor=darkblue}

\title{\textsc{Instruct-FD}: Can Your Full-Duplex Speech System Follow Turn-Taking Instructions?}

\author{\normalsize\bf Yuzhi Tang~~,~~ Wentao Ma~~,~~ Xiling Zhao~~,~~ Ahmad Salimi~~,~~ Sepehr Harfi Moridani,\\
\normalsize\bf Dongming Shen~~,~~ Jixuan Wang~~,~~ Abdulrahman Abdulrazzag~~,~~ Murdock Aubry~~,\\
\normalsize\bf Yu-Hua Chen~~,~~ Daniel Lee~~,~~ Jaewon Lee~~,~~ Jonah Mackey~~,~~ Silin Meng~~,\\
\normalsize\bf Nicholas Stranges~~,~~ Chenxu Xiong~~,~~ Hao Yu~~,~~ Yi Zhu~~,~~ Mu Li~~,~~ Alex Smola\\[0.5\baselineskip]
\normalsize Boson AI, Santa Clara, CA 95054
}

\begin{document}

\ifcolmsubmission
\linenumbers
\fi

\maketitle
\begingroup
\renewcommand{\thefootnote}{}
\footnotetext{Correspondence to: \texttt{yuzhi.tang@boson.ai}.}
\endgroup

\begin{abstract}
Current full-duplex (FD) spoken dialogue systems can produce fluid interactions, yet it remains unclear whether they can \emph{adapt their turn-taking behavior when explicitly instructed}. This is critical for real-world deployment, where conversational policies vary across applications (e.g., proactive tutoring vs.\ passive counseling). We introduce \textsc{Instruct-FD}, an instruction-conditioned benchmark for evaluating controllable turn management in FD systems. To enable this, we develop a human-validated, scalable synthetic pipeline that generates instruction-conditioned conversations, along with a deployment-agnostic multi-turn evaluation protocol and an LLM-based judge. Benchmarking six state-of-the-art full-duplex systems reveals a substantial gap in instruction-following turn management: the best model achieves only 64.4\% adherence. Performance is highly uneven across behaviors and scenarios, with proactive behaviors such as model backchanneling and interruption remaining particularly challenging. These findings establish instruction-following turn management as a crucial direction for building adaptable and deployable full-duplex dialogue systems.
\end{abstract}

\section{Introduction}

Full-duplex (FD) spoken dialogue systems~\citep{kyutai2024moshi, roy2026personaplexvoicerolecontrol, Google_Gemini_Live_2026, wang2024freezeomnismartlowlatency, yao2024minicpm}, which can listen and speak simultaneously---are a key step toward natural spoken interaction. Unlike strict turn-based systems, FD models must manage the conversational floor in real time: when to keep listening, when to backchannel, when to interrupt, when to continue speaking, and when to yield. These decisions are central to a deployed system, for instance: (1) A tutoring agent should interrupt early when a student makes a factual error. (2) A counseling agent, by contrast, should listen conservatively with minimal interruption. Both behaviors are valid, what differs is the turn-management policy. If a model cannot reliably adjust its turn-taking behavior in response to an explicit conversational instruction, deploying it across such diverse applications becomes impractical. A central evaluation need is therefore \emph{instruction-following turn management: can a model change how it takes turns when told to?}

Recent FD benchmarks~\citep{lin2025fullduplexbenchbenchmarkevaluatefullduplex, lin2026fullduplexbenchv15evaluatingoverlap, arora2025talkingturnsbenchmarkingaudio, zhang2026mtrduplexbenchcomprehensiveevaluationmultiround, lin2025fullduplexbenchv2multiturnevaluationframework} evaluate standard turn-taking quality but leave under-standardized the cases where the model must decide whether to speak \emph{during} user speech: existing tests tie backchannel and interruption quality to fixed priors or narrow safety scenarios~\citep{ge2025flexibenchmarkingfullduplexhumanllm}, and none test compliance with \emph{arbitrary} turn-taking instructions. We therefore frame turn-taking evaluation as an instruction-following problem and address two further gaps: (i)~\textbf{data scarcity}—we build a synthetic pipeline that generates policy-conditioned FD conversations with controllable turn triggers; (ii)~\textbf{context and portability}—existing multi-turn benchmarks~\citep{zhang2026mtrduplexbenchcomprehensiveevaluationmultiround, arora2025talkingturnsbenchmarkingaudio, lin2025fullduplexbenchv2multiturnevaluationframework} rely on dual-stream interfaces that limit deployability, so we pair contextual two-turn test cases with a multi-turn orchestrator streaming user audio over a unified WebRTC-compatible interface for scalable, architecture-agnostic evaluation.

\begin{figure}[htbp]
  \centering
  \includegraphics[width=\linewidth]{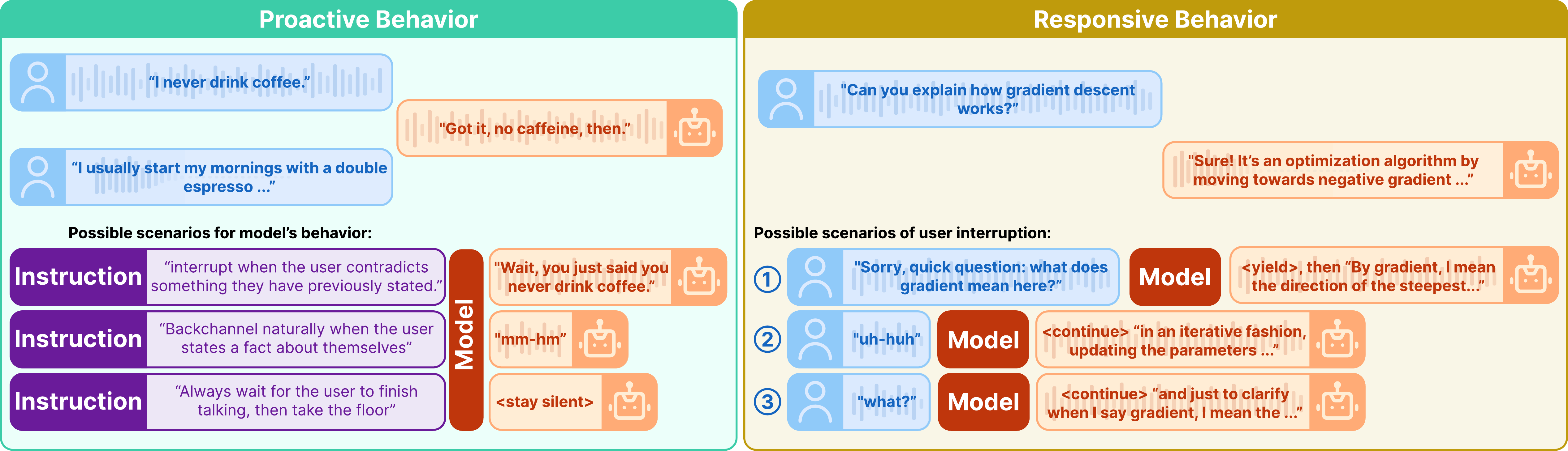}
  \caption{Overview of the \textsc{Instruct-FD} evaluation protocol. Left is the proactive behaviours, and right is the responsive behaviours.}
  \label{fig:eval-protocol}
\end{figure}

We present \textsc{Instruct-FD}, a benchmark that pairs user audio streams affording specific turn-management opportunities with explicit natural-language instructions. We evaluate two complementary situations: (1) when the user is speaking (should the model remain silent, give a short acknowledgment, or interrupt?) and (2) when the model is speaking and the user overlaps (should the model continue or acknowledge and adapt?). We show the general evaluation protocol in Figure~\ref{fig:eval-protocol}. This design focuses evaluation on controllable conversational behavior rather than imitation of a single timing distribution, and scales to 29 conversation scenarios with an extensible generation schema. Our contributions are:
\begin{itemize}
  \item \textbf{Instruction-conditioned evaluation framework.} We formulate turn-management evaluation as instruction following, explicitly covering both proactive behaviors (model backchanneling/interrupting during user speech) and responsive behaviors. We use a WebRTC-compatible multi-turn user orchestrator for architecture-agnostic multi-turn evaluation and LLM-as-a-judge for instruction adherence evaluation.
  \item \textbf{Synthetic instruction-following test-case generation pipeline.} We introduce a scalable pipeline for constructing instruction-conditioned FD conversations with broad and extensible coverage (29 scenarios), including contextual two-turn cases where turn-taking decisions depend on prior dialogue context; human validation confirms that generated cases are natural and instructions actionable.
  \item \textbf{Comprehensive model benchmarking and case-study diagnosis.} We benchmark six state-of-the-art full-duplex systems on \textsc{Instruct-FD} and use targeted case studies to identify current limitations, showing that instruction-following turn-management control remains a major open challenge.
\end{itemize}


\section{\textsc{Instruct-FD}}
\textit{For detailed related work, check Appendix~\ref{relatedwork}.} 
We now describe \textsc{Instruct-FD}, an evaluation framework that treats turn management in full-duplex spoken dialogue as an instruction-following problem. Given a set of turn-taking policies expressed as natural-language instructions, the framework generates targeted test scenarios, executes them against live models, and judges whether the model's behavior conforms to the specified policy.

\subsection{Turn-Management instruction taxonomy}
\label{subsec:turn-Management Instruction Taxonomy}

\begin{table}[htbp]
  \centering
  \scriptsize
  \setlength{\tabcolsep}{4pt}
  \begin{tabular}{@{}p{0.10\textwidth}p{0.10\textwidth}p{0.25\textwidth}p{0.5\textwidth}@{}}
    \toprule
    \textbf{Category} & \textbf{Instruction} & \textbf{Instruction Template} & \textbf{Example Instruction} \\
    \midrule
    Proactive & Backchannel & Backchannel when \textbf{\{trigger\}}, listen otherwise. & "You prefer to \textbf{backchannel briefly when the user is still speaking but appears to seek confirmation} (e.g., short pauses, upward intonation, emotional emphasis). Outside of those moments, do not interrupt and continue listening." \\[2pt]
    & Interrupt & Interrupt when \textbf{\{trigger\}}, listen otherwise & "\textbf{When the user makes a statement that directly contradicts something they said earlier in the conversation, interrupt immediately} to flag the inconsistency..." \\[2pt]
    & Listen & Listen, do not backchannel or interrupt & "You prefer to \textbf{listen silently while the user is speaking}. Do not interrupt or backchannel, and only take the floor after the user clearly finishes their turn." \\[2pt]
    \midrule
    Responsive & Continue & Continue when user backchannels & "While you are speaking, if the user gives a short acknowledgment or continuer (e.g., ``mm-hmm,'' ``yeah,'' ``got it''), treat it as support rather than a request to stop and \textbf{continue your response without yielding}." \\[2pt]
    & Acknowledge & Acknowledge when user interrupts & "While you are speaking, if the user interjects with a correction, constraint, clarification, or redirect, \textbf{acknowledge it immediately and adjust your response accordingly}." \\[2pt]
    \bottomrule
  \end{tabular}
  \caption{Turn-management instruction taxonomy used in Instruct-FD.}
  \vspace{-2ex}
  \label{tab:instruction-taxonomy}
\end{table}

We organize turn-management evaluation by \emph{who currently holds the floor}, separating \textbf{proactive} behaviors (the user is speaking and the model decides whether/how to intervene) from \textbf{responsive} behaviors (the model is speaking and reacts to user overlap). Inspired by prior FD benchmark taxonomies, especially FullDuplexBench v1 \citep{lin2025fullduplexbenchbenchmarkevaluatefullduplex}, we further condition each behavior on explicit policy instructions (Table~\ref{tab:instruction-taxonomy}). This design evaluates fundamental, controllable turn-taking abilities rather than assuming a single fixed interaction style, and enables same-conversation different-instruction comparisons that isolate instruction adherence from dialogue content. We describe each instruction category below:

\textbf{(1) Backchannel}: the model provides brief acknowledgments during user speech at appropriate moments (e.g., pauses, confirmation cues) without taking the floor.
\textbf{(2) Interrupt}: the model proactively takes the floor when an explicit trigger is present (e.g., contradiction, safety-critical signal), with both timely entry and functionally relevant content.
\textbf{(3) Listen}: the model remains silent during user speech, suppressing backchannels and interruptions until clear turn completion.
\textbf{(4) Continue}: when the user produces a short continuer (e.g., ``mm-hmm'') during model speech, the model maintains the floor and completes its thought rather than yielding.
\textbf{(5) Acknowledge}: when the user barges in with a correction or redirect during model speech, the model recognizes the interruption and adapts accordingly.

\subsection{Test case generation pipeline}
\label{subsec:test-case-generation}

\begin{figure}[htbp]
  \centering
  \includegraphics[width=\linewidth]{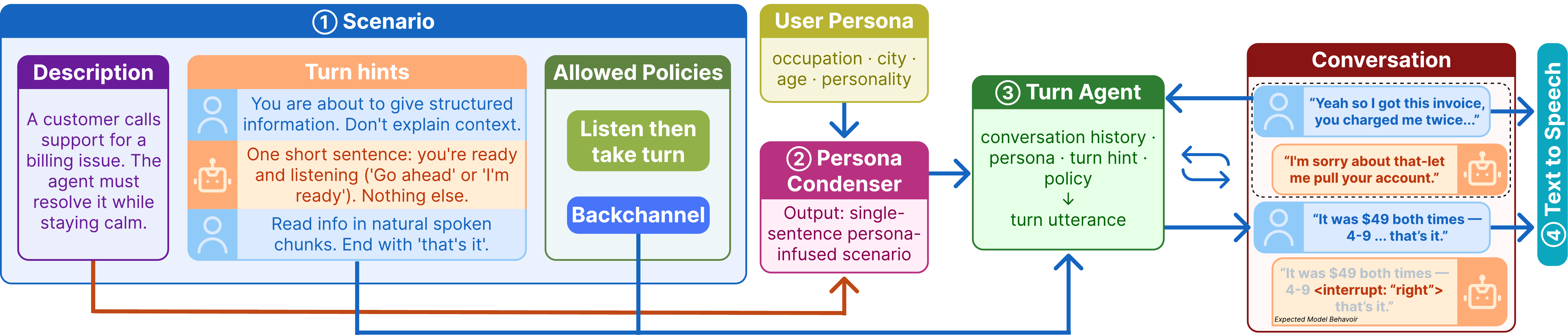}
  \caption{The four steps test case generation pipeline.}
  \label{fig:data-generation}
  \vspace{-2ex}
\end{figure}

Here we explain how we generate synthetic multi-turn conversations in which overlap events are explicitly scripted via inline transcript markers, enabling exact recovery of ground-truth insertion timestamps after TTS synthesis. As illustrated in Figure~\ref{fig:data-generation}, the pipeline proceeds in four stages.

\FloatBarrier

\textbf{(1) Text generation.} Each conversation is governed by a \emph{scenario specification} that defines the number of turns, speaker assignments, per-turn content hints, and an optional barge-in type. For persona-conditioned scenarios, a \emph{persona condenser} fuses one concrete attribute from a sampled Nemotron-Personas-USA profile~\citep{nvidia_persona} with the scenario description into a single grounding sentence. An LLM turn agent then generates the conversation turn by turn for both speakers, conditioning on the accumulated history and per-turn hints. For barge-in turns, the agent inserts structured inline markers (e.g., \texttt{<backchannel: PHRASE>}, \texttt{<interrupt: CONTENT>}) into the previous speaker's text at the intended cut-in point. Detailed templates and settings are provided in Appendix.

\textbf{(2) Speech synthesis and acoustic validation.} Each turn is synthesized with Gemini TTS, with barge-in sub-segments extracted from the inline markers and synthesized separately. ASR and forced alignment are then applied to recover word-level timestamps and locate the exact insertion point of each barge-in marker in the synthesized audio, yielding ground-truth tag timestamps. Cases with missing audio or alignment failures are discarded.

\textbf{(3) Test case assembly.} Transcript structure and acoustic metadata are merged into evaluation-ready test cases. A key design property is that the same conversation is paired with multiple turn-management instructions, each yielding a different valid target behavior. This same-conversation-different-instruction setup isolates instruction-following ability from conversational content variability. For each conversation, one test case is instantiated per instruction in the scenario's policy list, plus one un-instructed baseline.

\paragraph{(4) Balanced sampling.}
Balanced subsets are sampled across scenario types to ensure coverage. After filtering and sampling, the final \textsc{Instruct-FD} benchmark comprises 912 test cases across 29 scenarios. Table~\ref{tab:testcase-generation-breakdown} summarizes the distribution by instruction grouping. Notably, the gap between unique conversations and test case counts (e.g., 96 conversations yielding 312 test cases under \emph{Listen}) reflects the instruction-reuse design: a single conversation serves as the basis for multiple evaluation conditions.



\begin{table}[htbp]
  \centering
  \resizebox{0.85\columnwidth}{!}{%
    \begin{tabular}{lcccccc}
      \toprule
      & \multicolumn{3}{c}{\textbf{Proactive}} & \multicolumn{2}{c}{\textbf{Responsive}} & \textbf{Total} \\
      \cmidrule(lr){2-4} \cmidrule(lr){5-6} \cmidrule(lr){7-7}
      \textbf{Instruction} & \textbf{BC} & \textbf{Listen} & \textbf{Interrupt} & \textbf{Continue} & \textbf{Acknowledge} & \\
      \midrule
      \textbf{Scenarios} & 5 & 4 & 5 & 5 & 10 & 29  \\
      \textbf{Unique conversations} & 120 & 96 & 120 & 120 & 240 & 696  \\
      \textbf{Test cases} & 120 & 312 & 120 & 120 & 240 & 912  \\
      \bottomrule
    \end{tabular}
  }
  \caption{Distribution of scenarios and tests by instruction grouping.}
  \label{tab:testcase-generation-breakdown}
  \vspace{-2ex}
\end{table}

\subsection{Multi-turn user orchestrator}
\label{subsec:multi_turn_orchestrator}

\begin{figure}[htbp]
  \centering
  \includegraphics[width=0.8\linewidth]{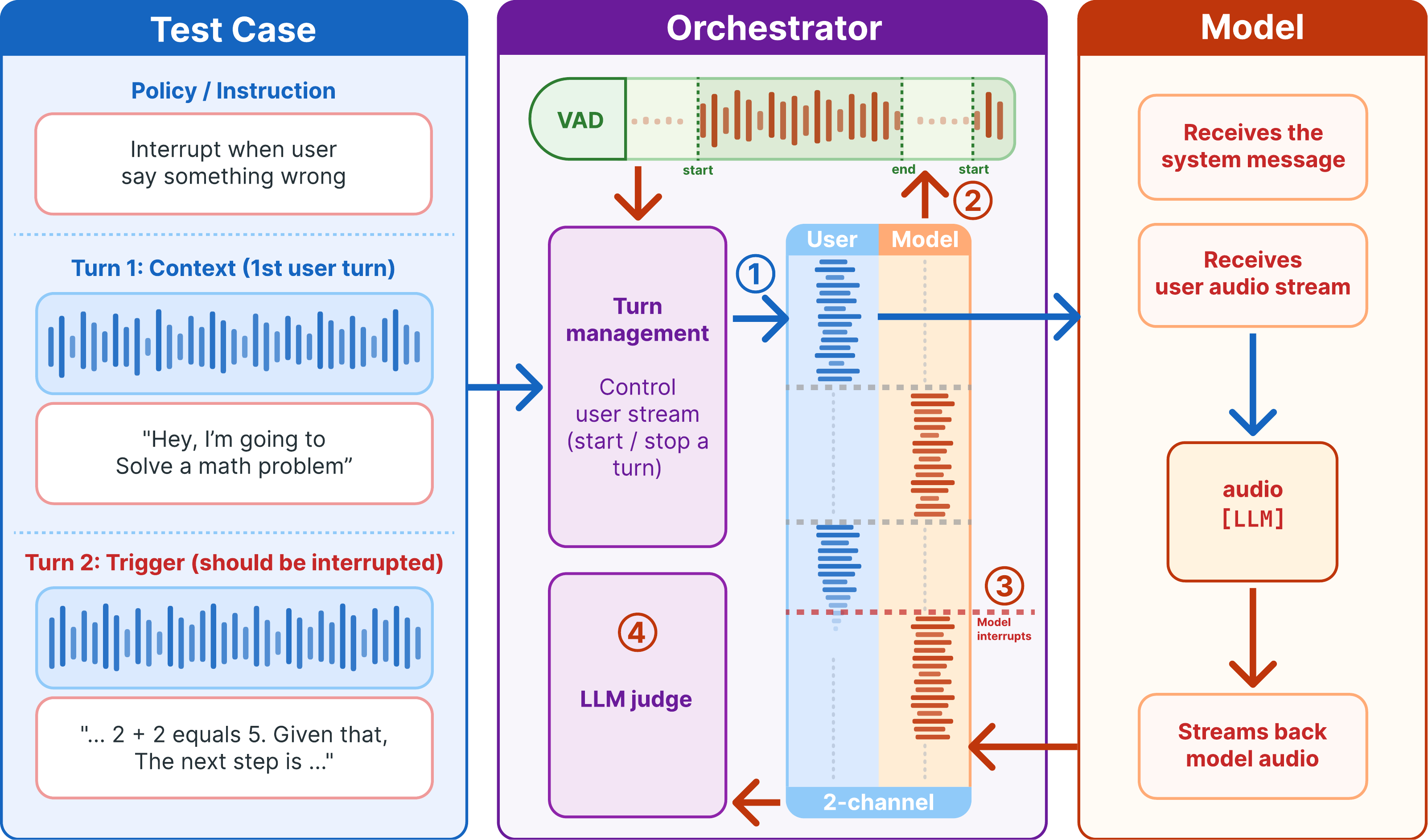}
  \caption{Multi-turn user orchestrator protocol. A turn manager streams scripted user audio via \texttt{take-turn} or \texttt{barge-in} actions; a VAD module governs turn boundaries and barge-in timing; user and model audio are logged into a time-aligned 2-channel WAV for judging.}
  \label{fig:orchestrator}
  \vspace{-2pt}
\end{figure}

To evaluate turn-management behavior in realistic multi-turn settings, we introduce a \textit{multi-turn user orchestrator} that controls the user-side audio stream while leaving model behavior fully unconstrained (Figure~\ref{fig:orchestrator}). Each test case contains (i) a system instruction encoding the turn-management policy and (ii) a turn manifest---a sequence of pre-recorded user audio segments with associated metadata. The orchestrator executes this manifest by streaming audio in real time within a single persistent model session, so model context accumulates across turns.

The orchestrator has three components.
\textbf{(1) Turn manager:} Each turn is one of two types. A \textit{take-turn} waits for VAD-detected model silence (default: 1.5\,s) or a per-turn timeout before streaming user audio. A \textit{barge-in} waits for the model to begin speaking, then injects user audio after a configurable delay relative to speech onset, simulating realistic user overlap. Trailing silence is appended after each user turn to reliably trigger the model's response. We show a demo test case schema in List~\ref{lst:test-case-schema}.
\textbf{(2) VAD module:} A Silero VAD~\citep{silero_vad} instance serves three roles: (a) detecting model silence to signal turn completion, (b) locating speech onset in user audio to compute barge-in timing, and (c) tracking whether the model is actively speaking during user audio streaming, which governs the barge-in handling policy---\textit{hold} (continue streaming regardless) or \textit{yield} (stop streaming and cede the floor).
\textbf{(3) Two-channel recording:} User and model audio chunks are timestamped against a shared session clock and accumulated in parallel logs. After each episode, the logs are rendered into a time-aligned 2-channel WAV (ch0\,=\,user, ch1\,=\,model), which serves as the artifact for downstream analysis and judging (\S\ref{subsec:llm_judge}).

Context turns (type \textit{take-turn}) first establish conversational state through normal exchange; trigger turns then expose policy-relevant overlap cues at controlled points, and the resulting joint audio is captured on the shared timeline. Because the orchestrator only controls the user channel and never injects into the model's response stream, every model is evaluated with identical streaming logic, timing structure, and user content, ensuring reproducibility and fair comparison across heterogeneous systems without requiring access to model internals.

\subsection{LLM-as-Judge for instruction adherence evaluation}
\label{subsec:llm_judge}

We evaluate model adherence to turn-management instructions using an LLM judge. We obtain word-level timestamps for both user and model audio using Qwen3-ASR-1.7B and Qwen3-ForcedAligner-0.6B. We then construct a unified transcript by interleaving the two timestamps and marking overlap regions. This yields a temporally grounded transcript that enables reasoning about both what was said and when it occurred.

For each case, the judge is provided with three inputs: 
(1) the instruction the model is expected to follow, 
(2) the reconstructed transcript, and 
(3) a groundtruth reference derived from the test case turn logs, encoding the expected turn-taking pattern.
We use Claude Sonnet 4.6 as the judge, and it follows a structured chain-of-thought process to (i) formalize instructions into rules, (ii) construct an event timeline of model behaviors, (iii) evaluate rule adherence under precise timing constraints, and (iv) produce a final verdict (example see Appendix~\ref{app:llm-judge-example}).
We validate the judge against human annotations and observe high agreement. A discussion appears in Section~\ref{sec:analysis}, and detailed error analysis is provided in Appendix~\ref{app:llm-judge-validation}.

Based on the judge outputs, we define two metrics.

\paragraph{Instruction Adherence Score (IAS).}
For a model \(m\) and instruction \(I\), we define IAS as the mean pass rate of the judge over test cases associated with \(I\):
\begin{equation}
\text{IAS}(m, I) = \frac{1}{|\mathcal{D}_I|} \sum_{x \in \mathcal{D}_I} \mathbf{1}\!\left[\text{judge}(x^{(I)}, I) = \text{pass}\right],
\end{equation}
where \(\mathcal{D}_I\) is the set of test cases associated with instruction \(I\), \(x^{(I)}\) is the model response for case \(x\) under instruction \(I\), and \(\mathbf{1}[\cdot]\) is the indicator function.

\paragraph{Instruction Sensitivity Score (\(\Delta\)IAS).}
To isolate the effect of instruction conditioning, we define a no-instruction counterpart on the same test cases:
\begin{equation}
\text{IAS}(m, \varnothing \mid I) = \frac{1}{|\mathcal{D}_I|} \sum_{x \in \mathcal{D}_I} \mathbf{1}\!\left[\text{judge}(x^{(\varnothing)}, I) = \text{pass}\right],
\end{equation}
where \(x^{(\varnothing)}\) is the model response for the same test case, generated without providing the instruction to the model. Both $x^{(\varnothing)}$ and $x^{(I)}$ are then evaluated by the judge against the same instruction $I$ on the same test set $\mathcal{D}_I$.

We then define:
\begin{equation}
\Delta \text{IAS}(m, I) = \text{IAS}(m, I) - \text{IAS}(m, \varnothing \mid I).
\end{equation}

IAS measures absolute adherence to the specified instruction, while \(\Delta\)IAS isolates the behavior change induced by instruction conditioning. A high IAS with low \(\Delta\)IAS suggests performance is largely driven by the model's inherent turn-taking behavior, whereas a high \(\Delta\)IAS indicates strong responsiveness to instruction.

\section{Experiments}
\subsection{Model evaluation setup}

We evaluate six full-duplex dialogue models using our instruction-conditioned test cases:
Freeze-Omni~\citep{wang2024freezeomnismartlowlatency}, Fun-Audio-Chat~\citep{funaudiochat2025}, Gemini-Live~\citep{Google_Gemini_Live_2026}, MiniCPM-o~\citep{yao2024minicpm}, Moshi~\citep{kyutai2024moshi}, and PersonaPlex~\citep{roy2026personaplexvoicerolecontrol}. We exclude GPT-Realtime~\citep{openai_chatgpt_realtime_2026} from our evaluation because its API requires the user to explicitly signal an interruption rather than allowing the model to autonomously manage the conversational floor, which is incompatible with our goal.

Each model is evaluated under all test cases in \textsc{Instruct-FD}, following each model's officially provided full-duplex streaming configuration as closely as possible while preserving default inference parameters. Where model-specific adjustments are necessary for fair evaluation, we document these in Appendix~\ref{app:inference-settings}. In \emph{barge-in} turns, user audio is injected 3\,s after the model begins speaking, simulating overlapping speech. Model responses are recorded as stereo WAV files (channel\,0: user, channel\,1: model) for downstream evaluation. API-based models (Gemini) run without local GPUs; open models run multiple parallel instances per node on 8$\times$H100 80\,GB GPUs via Slurm array jobs.

\subsection{Results}
\begin{table}[htbp]
  \centering
  \small
  \setlength{\tabcolsep}{3.8pt}
  \begin{tabular}{lcccccc}
    \toprule
    \textbf{Model} & \textbf{Overall} & \textbf{Listen} & \textbf{Backchannel} & \textbf{Interrupt} & \textbf{Continue} & \textbf{Acknowledge} \\
    \midrule
    Freeze-Omni & 10.7\% & 7.1\% & 0.0\% & 5.0\% & 40.0\% & 9.6\% \\
    Fun-Audio-Chat & 46.2\% & 90.7\% & 0.0\% & 9.2\% & 94.2\% & 5.4\% \\
    Gemini & 64.4\% & 72.4\% & 50.0\% & 5.8\% & 71.7\% & 86.3\% \\
    MiniCPM-o & 36.4\% & 59.6\% & 12.5\% & 1.7\% & 97.5\% & 5.4\% \\
    Moshi & 21.1\% & 12.8\% & 1.7\% & 11.7\% & 88.3\% & 12.0\% \\
    PersonaPlex & 47.0\% & 66.7\% & 5.8\% & 10.0\% & 90.0\% & 35.8\% \\
    \bottomrule
  \end{tabular}
  \caption{Instruction Adherence Score (\(\text{IAS}\)) by instruction and model. \textbf{Overall} reports IAS aggregated across all instruction-conditioned test cases.}
  \label{tab:benchmark-if-eval}
\end{table}

Table~\ref{tab:benchmark-if-eval} reports IAS across the six models. Gemini achieves the highest overall score (64.4\%), followed by PersonaPlex (47.0\%) and Fun-Audio-Chat (46.2\%). Performance is strongly instruction-dependent: Continue is high for most models (40.0\%--97.5\%), while Backchannel (0.0\%--50.0\%) and Interrupt (1.7\%--11.7\%) remain low across all models. The strongest per-instruction results are Fun-Audio-Chat on Listen (90.7\%), Gemini on Backchannel (50.0\%) and Acknowledge (86.3\%), Moshi on Interrupt (11.7\%), and MiniCPM-o on Continue (97.5\%).

\begin{table}[htbp]
  \centering
  \small
  \setlength{\tabcolsep}{4.0pt}
  \begin{tabular}{lccccc}
    \toprule
    \textbf{Model} & \textbf{$\Delta$ Listen} & \textbf{$\Delta$ Backchannel} & \textbf{$\Delta$ Interrupt} & \textbf{$\Delta$ Continue} & \textbf{$\Delta$ Acknowledge} \\
    \midrule
    Freeze-Omni    & $-1.0\%$ & $+0.0\%$ & $-2.5\%$ & $-0.8\%$ & $+0.8\%$ \\
    Fun-Audio-Chat & $-2.2\%$ & $+0.0\%$ & $\mathbf{+9.2\%}$ & $+0.0\%$ & $\mathbf{-11.2\%}$ \\
    Gemini         & $+3.5\%$ & $\mathbf{+42.5\%}$ & $+0.0\%$ & $+6.7\%$ & $\mathbf{+11.7\%}$ \\
    MiniCPM        & $\mathbf{+10.6\%}$ & $+6.7\%$ & $-5.0\%$ & $-0.8\%$ & $+3.3\%$ \\
    Moshi          & $+1.9\%$ & $-6.7\%$ & $+2.5\%$ & $+5.0\%$ & $+1.7\%$ \\
    PersonaPlex    & $+5.8\%$ & $-2.5\%$ & $+5.0\%$ & $+2.5\%$ & $+4.6\%$ \\
    \bottomrule
  \end{tabular}
  \caption{Instruction Sensitivity Score (\(\Delta\)IAS) by instruction and model. Bold indicates statistical significance (McNemar test, Holm-corrected \(p<0.05\)).}
  \label{tab:primary-policy-scores}
\end{table}

Table~\ref{tab:primary-policy-scores} reports \(\Delta\)IAS across the six models. Significant positive shifts are concentrated in a few settings: Gemini improves on Backchannel (+42.5\%) and Acknowledge (+11.7\%), MiniCPM-o improves on Listen (+10.6\%), and Fun-Audio-Chat improves on Interrupt (+9.2\%). A significant negative shift is also observed for Fun-Audio-Chat on Acknowledge ($-11.2\%$). Most remaining deltas are small in magnitude. 


\begin{table*}[htbp]
  \centering
  \scriptsize
  \setlength{\tabcolsep}{3pt}
  \begin{minipage}[t]{0.49\textwidth}
    \centering
    \begin{tabular}{@{}p{0.52\linewidth}ccc@{}}
      \toprule
      \textbf{Scenario} & \textbf{Interrupt} & \shortstack{\textbf{Back-}\\\textbf{channel}} & \textbf{Listen} \\
      \midrule
      Sequential Info Capture & - & - & 33.3\% \\
      User Attention Check & - & 100.0\% & - \\
      Clause Boundary Tracking & - & 0.0\% & 95.8\% \\
      User Filler Pause & - & 41.7\% & 62.5\% \\
      Emotional Disclosure & - & 16.7\% & 79.2\% \\
      Floor Hold Signal & - & 91.7\% & 8.3\% \\
      Factual Misinformation & 0.0\% & - & 100.0\% \\
      Hesitation Prompt & 25.0\% & - & 75.0\% \\
      Safety Correction & 0.0\% & - & 95.8\% \\
      Self Contradiction & 4.2\% & - & 91.7\% \\
      Word Retrieval Assist & 0.0\% & - & 100.0\% \\
      Cognitive Pause & - & - & 83.3\% \\
      User Projection Pause & - & - & 83.3\% \\
      Self Directed Question & - & - & 33.3\% \\
      \bottomrule
    \end{tabular}
    \caption{Scenario-level IAS for Gemini on proactive instructions.}
    \label{tab:gemini-scenario-main-proactive}
  \end{minipage}
  \hfill
  \begin{minipage}[t]{0.49\textwidth}
    \centering
    \begin{tabular}{@{}p{0.52\linewidth}cc@{}}
      \toprule
      \textbf{Scenario} & \textbf{Continue} & \textbf{Acknowledge} \\
      \midrule
      User Acknowledgment & 70.8\% & - \\
      User Alignment & 58.3\% & - \\
      User Amusement & 70.8\% & - \\
      User Understanding & 62.5\% & - \\
      User Continuer & 95.8\% & - \\
      Short Stop Repair & - & 91.7\% \\
      Short Repeat Request & - & 95.8\% \\
      Short Confusion Signal & - & 70.8\% \\
      Hearing Check & - & 79.2\% \\
      Add Constraint & - & 100.0\% \\
      Topic Redirect & - & 87.5\% \\
      Scope Narrowing & - & 95.8\% \\
      Skip Basics & - & 66.7\% \\
      Summary Request & - & 87.5\% \\
      Personal Context & - & 91.7\% \\
      \bottomrule
    \end{tabular}
    \caption{Scenario-level IAS for Gemini on responsive instructions.}
    \label{tab:gemini-scenario-main-responsive}
  \end{minipage}
\end{table*}

Here we also show the per-scenario IAS for Gemini (Tables~\ref{tab:gemini-scenario-main-proactive} and~\ref{tab:gemini-scenario-main-responsive}): they show substantial within-instruction variation. In proactive Backchannel scenarios, IAS range from 0.0\% (Clause Boundary Tracking) to 100.0\% (User Attention Check). In proactive Interrupt scenarios, IAS remain low (0.0\%--25.0\%). In responsive scenarios, Continue is strongest for User Continuer (95.8\%) and lower for User Alignment (58.3\%), while Acknowledge ranges from 66.7\% (Skip Basics) to 100.0\% (Add Constraint).

\section{Analysis}
\label{sec:analysis}
\textbf{\textit{Is the LLM judge reliable?}} We assess the reliability of the LLM judge using 180 manually annotated judgments across models and instructions. The judge achieves 88.9\% accuracy (87.6 macro-F1). A qualitative analysis of error cases shows that most judge errors are associated with ASR transcript artifacts and instruction misinterpretation.

\textbf{\textit{Will Semantic compliance sacrifice timing compliance?}} A recurring failure mode is semantic compliance without temporal compliance: models often produce content that is semantically appropriate to the policy, but at an incorrect timing point. For example, instead of backchanneling during user speech, the model waits and produces a valid acknowledgment after the user has finished, which no longer satisfies a proactive criterion. This explains why responses can sound cooperative in qualitative inspection while still failing strict instruction-following metrics for Backchannel and Interrupt.

\textbf{\textit{Where do models fail?}} To better understand the common failures, we highlight representative patterns that recur across models and scenarios:

\textit{1) Trigger hierarchy in backchanneling (Gemini-Live):}
Gemini’s backchannel behavior appears strongly scenario-dependent (Table~\ref{tab:gemini-scenario-main-proactive}). Performance is highest on explicit cues such as \textit{User Attention Check} (100.0\%) and \textit{Floor Hold Signal} (91.7\%), where the user explicitly checks in with the model and briefly pauses. In contrast, performance drops on more subtle signals, including \textit{User Filler Pause} (41.7\%), \textit{Emotional Disclosure} (16.7\%), and \textit{Clause Boundary Tracking} (0.0\%). This suggests that Gemini can follow backchannel instructions when triggers are salient, but struggles with weaker, less explicit conversational cues. See Appendix~\ref{app:gemini-backchannel-case} for example model responses.

\textit{2) Continue default dominates responsive behavior (Freeze-Omni, Fun-Audio-Chat, MiniCPM-o):}
Across these models, responsive overlap handling often collapses into a continue-default policy. Fun-Audio-Chat and MiniCPM-o achieve high Continue scores in many responsive scenarios (often near 100\%) while Acknowledge remains low in repair-like cases (Tables~\ref{tab:funaudiochat-scenario-appendix-responsive} and~\ref{tab:minicpm-scenario-appendix-responsive}). Freeze-Omni shows the same direction at lower absolute performance, with Continue generally exceeding Acknowledge (Table~\ref{tab:freezeomni-scenario-appendix-responsive}). This pattern indicates weak discrimination between ``user continuer'' overlaps and ``user redirect/repair'' overlaps.

\textit{3) Architecture vs. training data effect (Moshi vs PersonaPlex):}
PersonaPlex improves substantially over Moshi on overall instruction following (47.0\% vs.\ 21.1\%), Listen (66.7\% vs.\ 12.8\%), and Acknowledge (35.8\% vs.\ 12.0\%) in Table~\ref{tab:benchmark-if-eval}, suggesting that additional training beyond the base architecture improves baseline turn management. However, both models remain weak on proactive speaking actions and show limited instruction-induced shifts (Table~\ref{tab:primary-policy-scores}), indicating that better baseline interaction quality does not automatically translate into robust instruction controllability. See Appendix~\ref{app:moshi-personaplex-turn-taking} for a qualitative side-by-side transcript case.

\textit{4) Instructions destabilize acknowledge behavior (Fun-Audio-Chat):}
Fun-Audio-Chat exhibits a notable negative instruction effect: Acknowledge drops significantly under instruction ($-11.2\%$), even as Interrupt increases ($+9.2\%$) (Table~\ref{tab:primary-policy-scores}). Combined with near-zero Acknowledge performance in many responsive repair scenarios (Table~\ref{tab:funaudiochat-scenario-appendix-responsive}), this suggests that stronger instruction prompting may destabilize behavior when the model lacks fine-grained turn-management control. See Appendix~\ref{app:negative-instruction-funaudiochat} for a paired qualitative example.

\section{Human study for benchmark validation}

We conduct a human study to validate two properties of the benchmark: the naturalness of the test cases and the actionability of the turn-management instructions. Specifically, we ask two questions: (1) can humans follow the test case instruction and identify appropriate action timings in each case (i.e., when to backchannel, interrupt, or listen, and whether to continue or acknowledge after a user interjection), and (2) how sensitive is human behavior to different turn-management instructions?

For proactive instruction cases, we play the user--model conversation audio from the test case turn logs up to the user turn containing the trigger. Participants are asked to decide turn-taking action timing for the model given instructions. During that user turn, participants are prompted to make a decision at approximately 2s intervals: whether the model should Backchannel, Interrupt, or Listen under given instruction. For responsive instruction cases, we play the conversation through the user interjection overlap (up to 2 seconds) and ask participants whether the model should Continue or Acknowledge. We compare participant responses against ground-truth labels derived from test-case metadata to compute human action-timing accuracy. We also evaluate each case without providing the instruction to the participant, to estimate human default turn-taking behavior and compute the instruction-induced change in case-level accuracy.

We recruit 11 participants fluent in English. Each participant is assigned 114--115 test cases, with no repeated conversation for the same participant to avoid memorization. Assignments are balanced across instructions and scenarios. In total, participants complete 1,199 cases with 4,031 decision points.

\paragraph{Results.}
Participants achieved \textbf{89.6\%} case accuracy (84.1\% proactive, 98.5\% responsive), with per-scenario breakdowns in Tables~\ref{tab:scenario-proactive-placeholder} and~\ref{tab:scenario-responsive-placeholder}. The consistently high accuracy across 29 scenarios categories provides evidence that our synthetic test cases present natural conversational situations in which the correct turn-management action is clearly identifiable. Backchannel is the most challenging action (63.7\%), reflecting a compound criterion and a known annotation mismatch between offline LLM-selected timestamps and online human perception; we provide explanations for this in Appendix~\ref{app:backchannel-analysis}. Table ~\ref{tab:human_primary-policy-scores} shows giving participants instructions produce significant behavioral shifts for Listen ($\Delta = +38.6\%$, $p < 0.001$) and Interrupt ($\Delta = +36.5\%$, $p < 0.001$) cases. Behavioral shift for Backchannel, Acknowledge, and Continue deltas are also positive but not statistically significant ($p > 0.37$). Overall, these results indicate that the benchmark presents natural conversational scenarios, and that participants can reliably follow turn-management instructions. Moreover, the instruction-induced behavioral shifts demonstrate that the instructions are actionable and can meaningfully influence human decision-making, supporting the validity of our evaluation framework.




\section{Conclusion and Future Work}

In this work, we presented \textsc{Instruct-FD}, an instruction-conditioned benchmark for evaluating turn-management behavior in full-duplex spoken dialogue. By framing turn-taking as an instruction-following problem, our framework enables controlled evaluation of five instruction types across 29 scenarios using a synthetic generation pipeline, a deployment-agnostic multi-turn orchestrator, and an LLM judge validated against human annotations. Benchmarking six systems reveals that controllable turn-taking remains a key bottleneck: the best model reaches 64.4\% overall instruction adherence, with proactive behaviors such as Backchannel and Interrupt substantially lagging behind responsive policies, and instruction sensitivity varying widely across models. Beyond aggregate metrics, our analysis surfaces systematic failure patterns—trigger-dependent backchannel performance, continue-default collapse in responsive overlap, and instruction-induced destabilization—that point to concrete directions for model improvement. A human study on the same test cases confirms that the scenarios are interpretable and the instructions are actionable.

Several directions remain open for future work. First, we evaluate a fixed set of five policy instructions chosen for broad coverage of proactive and responsive turn-management behaviors. The instruction-conditioned evaluation protocol itself is not limited to these five; extending to richer free-form instructions, compositional instructions, and finer-grained behavioral variants is a natural next step. Second, all scenarios are constrained to interactions of around two model turns. This is not an oversight but a principled design choice: because user-side audio must be pre-generated before evaluation, longer dialogues would require user continuations that depend on hypothetical model responses, introducing artificial branch-selection artifacts that confound turn-management measurement. Overcoming this constraint is an open research problem; promising directions include interactive test generation with branching dialogue trees, or hybrid protocols that combine pre-scripted context turns with live user-model interaction at trigger points. Third, our benchmark currently covers English only; cross-lingual extension would be valuable given that turn-taking conventions vary substantially across languages and cultures.

\clearpage
\bibliography{colm2026_conference}

\begin{thebibliography}{31}
\providecommand{\natexlab}[1]{#1}
\providecommand{\url}[1]{\texttt{#1}}
\expandafter\ifx\csname urlstyle\endcsname\relax
  \providecommand{\doi}[1]{doi: #1}\else
  \providecommand{\doi}{doi: \begingroup \urlstyle{rm}\Url}\fi

\bibitem[Ardila et~al.(2020)Ardila, Branson, Davis, Henretty, Kohler, Meyer, Morais, Saunders, Tyers, and Weber]{ardila2020commonvoice}
Rosana Ardila, Megan Branson, Kelly Davis, Michael Henretty, Michael Kohler, Josh Meyer, Reuben Morais, Lindsay Saunders, Francis~M. Tyers, and Gregor Weber.
\newblock Common voice: A massively multilingual speech corpus.
\newblock In \emph{Proceedings of LREC}, 2020.
\newblock URL \url{https://arxiv.org/abs/1912.06670}.

\bibitem[Arora et~al.(2025)Arora, Lu, Chiu, Pang, and Watanabe]{arora2025talkingturnsbenchmarkingaudio}
Siddhant Arora, Zhiyun Lu, Chung-Cheng Chiu, Ruoming Pang, and Shinji Watanabe.
\newblock Talking turns: Benchmarking audio foundation models on turn-taking dynamics, 2025.
\newblock URL \url{https://arxiv.org/abs/2503.01174}.

\bibitem[Chen et~al.(2025)Chen, Cheng, Deng, Li, Liu, Tan, Wang, Xu, Ye, Zhang, Zhang, and Zhou]{funaudiochat2025}
Qian Chen, Luyao Cheng, Chong Deng, Xiangang Li, Jiaqing Liu, Chao-Hong Tan, Wen Wang, Junhao Xu, Jieping Ye, Qinglin Zhang, Qiquan Zhang, and Jingren Zhou.
\newblock Fun-audio-chat technical report.
\newblock 2025.
\newblock URL \url{https://arxiv.org/abs/2512.20156}.

\bibitem[Chen et~al.(2024)Chen, Yue, Zhang, Gao, Tan, and Li]{chen2024voicebench}
Yiming Chen, Xianghu Yue, Chen Zhang, Xiaoxue Gao, Robby~T. Tan, and Haizhou Li.
\newblock {VoiceBench: Benchmarking LLM‑Based Voice Assistants}.
\newblock \emph{arXiv preprint arXiv:2410.17196}, 2024.
\newblock URL \url{https://arxiv.org/abs/2410.17196}.

\bibitem[Chu et~al.(2023)Chu, Xu, Zhou, Yang, Zhang, Yan, Zhou, and Zhou]{qwen2023audio}
Yunfei Chu, Jin Xu, Xiaohuan Zhou, Qian Yang, Shiliang Zhang, Zhijie Yan, Chang Zhou, and Jingren Zhou.
\newblock {Qwen-Audio: Advancing Audio-Language Models}, 2023.
\newblock URL \url{https://arxiv.org/abs/2311.07919}.

\bibitem[D\'efossez et~al.(2024)D\'efossez, Mazar\'e, Orsini, Royer, P\'erez, J\'egou, Grave, and Zeghidour]{kyutai2024moshi}
Alexandre D\'efossez, Laurent Mazar\'e, Manu Orsini, Am\'elie Royer, Patrick P\'erez, Herv\'e J\'egou, Edouard Grave, and Neil Zeghidour.
\newblock Moshi: a speech-text foundation model for real-time dialogue.
\newblock Technical report, 2024.
\newblock URL \url{https://arxiv.org/abs/2410.00037}.

\bibitem[Ge et~al.(2025)Ge, Chen, Xiao, Liu, Xiao, Xiang, Yu, and Zhu]{ge2025flexibenchmarkingfullduplexhumanllm}
Yuan Ge, Saihan Chen, Jingqi Xiao, Xiaoqian Liu, Tong Xiao, Yan Xiang, Zhengtao Yu, and Jingbo Zhu.
\newblock Flexi: Benchmarking full-duplex human-llm speech interaction, 2025.
\newblock URL \url{https://arxiv.org/abs/2509.22243}.

\bibitem[Ghosh et~al.(2024)Ghosh, Kumar, Seth, Evuru, Tyagi, Sakshi, Nieto, Duraiswami, and Manocha]{ghosh2024gama}
Sreyan Ghosh, Sonal Kumar, Ashish Seth, Chandra Kiran~Reddy Evuru, Utkarsh Tyagi, S~Sakshi, Oriol Nieto, Ramani Duraiswami, and Dinesh Manocha.
\newblock {GAMA: A General Audio Model for Audio Understanding}.
\newblock \emph{arXiv preprint arXiv:2406.11768}, 2024.
\newblock URL \url{https://arxiv.org/abs/2406.11768}.

\bibitem[{Google}(2026)]{Google_Gemini_Live_2026}
{Google}.
\newblock Gemini live, 2026.
\newblock URL \url{https://gemini.google/overview/gemini-live/}.

\bibitem[Hernandez et~al.(2018)Hernandez, Nguyen, Ghannay, Tomashenko, and Est{\`e}ve]{hernandez2018tedlium3}
Fran{\c c}ois Hernandez, Vincent Nguyen, Sahar Ghannay, Natalia Tomashenko, and Yannick Est{\`e}ve.
\newblock {TED-LIUM 3: Twice as Much Data and Corpus Repartition for Experiments on Speaker Adaptation}.
\newblock In \emph{Proceedings of Interspeech}, 2018.
\newblock URL \url{https://arxiv.org/abs/1805.04699}.

\bibitem[Huang et~al.(2023)Huang, Li, Yang, Shi, Chang, Ye, Wu, Hong, Huang, Liu, Ren, Zhao, and Watanabe]{huang2023audiogpt}
Rongjie Huang, Mingze Li, Dongchao Yang, Jiatong Shi, Xuankai Chang, Zhenhui Ye, Yuning Wu, Zhiqing Hong, Jiawei Huang, Jinglin Liu, Yi~Ren, Zhou Zhao, and Shinji Watanabe.
\newblock {AudioGPT: Understanding and Generating Speech, Music, Sound, and Talking Head}.
\newblock \emph{arXiv preprint arXiv:2304.12995}, 2023.
\newblock URL \url{https://arxiv.org/abs/2304.12995}.

\bibitem[Lin et~al.(2025{\natexlab{a}})Lin, Kuan, Shi, Chang, Arora, Watanabe, and yi~Lee]{lin2025fullduplexbenchv2multiturnevaluationframework}
Guan-Ting Lin, Shih-Yun~Shan Kuan, Jiatong Shi, Kai-Wei Chang, Siddhant Arora, Shinji Watanabe, and Hung yi~Lee.
\newblock Full-duplex-bench-v2: A multi-turn evaluation framework for duplex dialogue systems with an automated examiner, 2025{\natexlab{a}}.
\newblock URL \url{https://arxiv.org/abs/2510.07838}.

\bibitem[Lin et~al.(2025{\natexlab{b}})Lin, Lian, Li, Wang, Anumanchipalli, Liu, and yi~Lee]{lin2025fullduplexbenchbenchmarkevaluatefullduplex}
Guan-Ting Lin, Jiachen Lian, Tingle Li, Qirui Wang, Gopala Anumanchipalli, Alexander~H. Liu, and Hung yi~Lee.
\newblock Full-duplex-bench: A benchmark to evaluate full-duplex spoken dialogue models on turn-taking capabilities, 2025{\natexlab{b}}.
\newblock URL \url{https://arxiv.org/abs/2503.04721}.

\bibitem[Lin et~al.(2026)Lin, Kuan, Wang, Lian, Li, Watanabe, and yi~Lee]{lin2026fullduplexbenchv15evaluatingoverlap}
Guan-Ting Lin, Shih-Yun~Shan Kuan, Qirui Wang, Jiachen Lian, Tingle Li, Shinji Watanabe, and Hung yi~Lee.
\newblock Full-duplex-bench v1.5: Evaluating overlap handling for full-duplex speech models, 2026.
\newblock URL \url{https://arxiv.org/abs/2507.23159}.

\bibitem[Meyer \& Corneil(2025)Meyer and Corneil]{nvidia_persona}
Yev Meyer and Dane Corneil.
\newblock {Nemotron-Personas-USA}: Synthetic personas aligned to real-world distributions, 2025.
\newblock URL \url{https://huggingface.co/datasets/nvidia/Nemotron-Personas-USA}.

\bibitem[Nguyen et~al.(2022)Nguyen, Kharitonov, Copet, Adi, Hsu, Elkahky, Tomasello, Algayres, Sagot, Mohamed, and Dupoux]{nguyen2022generativespokendialoguelanguage}
Tu~Anh Nguyen, Eugene Kharitonov, Jade Copet, Yossi Adi, Wei-Ning Hsu, Ali Elkahky, Paden Tomasello, Robin Algayres, Benoit Sagot, Abdelrahman Mohamed, and Emmanuel Dupoux.
\newblock Generative spoken dialogue language modeling, 2022.
\newblock URL \url{https://arxiv.org/abs/2203.16502}.

\bibitem[{OpenAI}(2026)]{openai_chatgpt_realtime_2026}
{OpenAI}.
\newblock Chatgpt realtime.
\newblock \url{https://platform.openai.com/docs/models/gpt-realtime}, 2026.

\bibitem[Panayotov et~al.(2015)Panayotov, Chen, Povey, and Khudanpur]{panayotov2015librispeech}
Vassil Panayotov, Guoguo Chen, Daniel Povey, and Sanjeev Khudanpur.
\newblock {LibriSpeech: An ASR Corpus Based on Public Domain Audio Books}.
\newblock In \emph{Proceedings of ICASSP}, 2015.
\newblock URL \url{https://ieeexplore.ieee.org/document/7178964}.

\bibitem[Pratap et~al.(2020)Pratap, Xu, Sriram, Synnaeve, and Collobert]{pratap2020mls}
Vineel Pratap, Qiantong Xu, Anuroop Sriram, Gabriel Synnaeve, and Ronan Collobert.
\newblock {MLS: A Large-Scale Multilingual Dataset for Speech Research}.
\newblock In \emph{Proceedings of Interspeech}, 2020.
\newblock URL \url{https://arxiv.org/abs/2012.03411}.

\bibitem[Roy et~al.(2026)Roy, Raiman, gil Lee, Ene, Kirby, Kim, Kim, and Catanzaro]{roy2026personaplexvoicerolecontrol}
Rajarshi Roy, Jonathan Raiman, Sang gil Lee, Teodor-Dumitru Ene, Robert Kirby, Sungwon Kim, Jaehyeon Kim, and Bryan Catanzaro.
\newblock Personaplex: Voice and role control for full duplex conversational speech models, 2026.
\newblock URL \url{https://arxiv.org/abs/2602.06053}.

\bibitem[Rubenstein et~al.(2023)Rubenstein, Asawaroengchai, Nguyen, Bapna, Borsos, {de Chaumont Quitry}, Chen, {El Badawy}, Han, Kharitonov, Muckenhirn, Padfield, Qin, Rozenberg, Sainath, Schalkwyk, Sharifi, Ramanovich, Tagliasacchi, Tudor, Velimirović, Vincent, Yu, Wang, Zayats, Zeghidour, Zhang, Zhang, Zilka, and Frank]{rubenstein2023audiopalm}
Paul~K. Rubenstein, Chulayuth Asawaroengchai, Duc~Dung Nguyen, Ankur Bapna, Zalán Borsos, Félix {de Chaumont Quitry}, Peter Chen, Dalia {El Badawy}, Wei Han, Eugene Kharitonov, Hannah Muckenhirn, Dirk Padfield, James Qin, Danny Rozenberg, Tara Sainath, Johan Schalkwyk, Matt Sharifi, Michelle~Tadmor Ramanovich, Marco Tagliasacchi, Alexandru Tudor, Mihajlo Velimirović, Damien Vincent, Jiahui Yu, Yongqiang Wang, Vicky Zayats, Neil Zeghidour, Yu~Zhang, Zhishuai Zhang, Lukas Zilka, and Christian Frank.
\newblock {AudioPaLM: A Large Language Model That Can Speak and Listen}.
\newblock \emph{arXiv preprint arXiv:2306.12925}, 2023.
\newblock URL \url{https://arxiv.org/abs/2306.12925}.

\bibitem[Sakshi et~al.(2024)Sakshi, Tyagi, Kumar, Seth, Selvakumar, Nieto, Duraiswami, Ghosh, and Manocha]{sakshi2024mmaumultitask}
S~Sakshi, Utkarsh Tyagi, Sonal Kumar, Ashish Seth, Ramaneswaran Selvakumar, Oriol Nieto, Ramani Duraiswami, Sreyan Ghosh, and Dinesh Manocha.
\newblock {MMAU: A Massive Multi‑Task Audio Understanding and Reasoning Benchmark}.
\newblock \emph{arXiv preprint arXiv:2410.19168}, 2024.
\newblock URL \url{https://arxiv.org/abs/2410.19168}.

\bibitem[Shah et~al.(2024)Shah, Noguero, Heikkila, Raj, and Kourtellis]{shah2024speech}
Muhammad~A Shah, David~Solans Noguero, Mikko~A Heikkila, Bhiksha Raj, and Nicolas Kourtellis.
\newblock {Speech robust bench: A robustness benchmark for speech recognition}.
\newblock \emph{arXiv preprint arXiv:2403.07937}, 2024.

\bibitem[{Silero Team}(2024)]{silero_vad}
{Silero Team}.
\newblock Silero vad: pre-trained enterprise-grade voice activity detector (vad), number detector and language classifier.
\newblock \url{https://github.com/snakers4/silero-vad}, 2024.

\bibitem[Tang et~al.(2024)Tang, Yu, Sun, Chen, Tan, Li, Lu, Ma, and Zhang]{sun2024salmonn}
Changli Tang, Wenyi Yu, Guangzhi Sun, Xianzhao Chen, Tian Tan, Wei Li, Lu~Lu, Zejun Ma, and Chao Zhang.
\newblock {SALMONN: Towards Generic Hearing Abilities for Large Language Models}.
\newblock In \emph{Proceedings of ICLR}, 2024.
\newblock URL \url{https://arxiv.org/abs/2310.13289}.

\bibitem[Tay et~al.(2026)Tay, Ma, Lee, Tang, Lee, Yin, Shen, Meng, Zhu, Li, et~al.]{tay2026back}
Geeyang Tay, Wentao Ma, Jaewon Lee, Yuzhi Tang, Daniel Lee, Weisu Yin, Dongming Shen, Silin Meng, Yi~Zhu, Mu~Li, et~al.
\newblock Back to basics: Revisiting asr in the age of voice agents.
\newblock \emph{arXiv preprint arXiv:2603.25727}, 2026.

\bibitem[Wang et~al.(2025{\natexlab{a}})Wang, Zou, Lin, Sun, Liu, Zhang, Liu, Aw, and Chen]{wang2025audiobench}
Bin Wang, Xunlong Zou, Geyu Lin, Shuo Sun, Zhuohan Liu, Wenyu Zhang, Zhengyuan Liu, AiTi Aw, and Nancy~F. Chen.
\newblock Audiobench: A universal benchmark for audio large language models.
\newblock In \emph{Proceedings of the 2025 Conference of the Nations of the Americas Chapter of the Association for Computational Linguistics: Human Language Technologies (NAACL)}, pp.\  4297--4316. Association for Computational Linguistics, 2025{\natexlab{a}}.

\bibitem[Wang et~al.(2025{\natexlab{b}})Wang, Ma, Guo, Wang, Xie, Xu, and Lin]{wang2025contextasr}
He~Wang, Linhan Ma, Dake Guo, Xiong Wang, Lei Xie, Jin Xu, and Junyang Lin.
\newblock {Contextasr-bench: A massive contextual speech recognition benchmark}.
\newblock \emph{arXiv preprint arXiv:2507.05727}, 2025{\natexlab{b}}.

\bibitem[Wang et~al.(2024)Wang, Li, Fu, Shen, Xie, Li, Sun, and Ma]{wang2024freezeomnismartlowlatency}
Xiong Wang, Yangze Li, Chaoyou Fu, Yunhang Shen, Lei Xie, Ke~Li, Xing Sun, and Long Ma.
\newblock Freeze-omni: A smart and low latency speech-to-speech dialogue model with frozen llm, 2024.
\newblock URL \url{https://arxiv.org/abs/2411.00774}.

\bibitem[Yao et~al.(2024)Yao, Yu, Zhang, Wang, Cui, Zhu, Cai, Li, Zhao, He, et~al.]{yao2024minicpm}
Yuan Yao, Tianyu Yu, Ao~Zhang, Chongyi Wang, Junbo Cui, Hongji Zhu, Tianchi Cai, Haoyu Li, Weilin Zhao, Zhihui He, et~al.
\newblock Minicpm-v: A gpt-4v level mllm on your phone.
\newblock \emph{arXiv preprint arXiv:2408.01800}, 2024.

\bibitem[Zhang et~al.(2026)Zhang, Cui, Xu, Li, Zhu, Bai, Ma, and King]{zhang2026mtrduplexbenchcomprehensiveevaluationmultiround}
He~Zhang, Wenqian Cui, Haoning Xu, Xiaohui Li, Lei Zhu, Haoli Bai, Shaohua Ma, and Irwin King.
\newblock Mtr-duplexbench: Towards a comprehensive evaluation of multi-round conversations for full-duplex speech language models, 2026.
\newblock URL \url{https://arxiv.org/abs/2511.10262}.

\end{thebibliography}
\bibliographystyle{colm2026_conference}
\clearpage
\appendix
\section{Appendix}

\subsection{Related work}
\label{relatedwork}
\subsubsection{Full-Duplex dialogue evaluation benchmarks}
Recent benchmarks have established evaluation frameworks beyond traditinal Automatic speech recognition (ASR)~\citep{panayotov2015librispeech,hernandez2018tedlium3,ardila2020commonvoice,tay2026back, pratap2020mls, shah2024speech, wang2025contextasr}, half-duplex (HD)~\citep{wang2025audiobench,chen2024voicebench,sakshi2024mmaumultitask} to full-duplex (FD) spoken dialogue systems along several complementary dimensions. 

One line of work focuses on \emph{(1) standard turn-taking quality}---smooth transitions, pause handling, and user interruption response. FullDuplexBench (FDB) v1~\citep{lin2025fullduplexbenchbenchmarkevaluatefullduplex} introduced a foundational taxonomy of these scenarios with reproducible automatic metrics on single user-stream audio, FDB v1.5~\citep{lin2026fullduplexbenchv15evaluatingoverlap} refined these protocols and added LLM-based judging for model behavior and response quality, and Taking-Turns~\citep{arora2025talkingturnsbenchmarkingaudio} uses a turn-taking prediction model trained on the Switchboard dataset to evaluate transition timing. 

A second line of work extends evaluation to \emph{(2) model-initiated behaviors during user speech}, such as backchanneling and proactive interruption. FLEXI~\citep{ge2025flexibenchmarkingfullduplexhumanllm} expanded coverage for model interruption, though its evaluation is scoped primarily to safety-centric cases. Backchannel quality in existing work is typically assessed via distribution matching against a fixed timing prior, which ties evaluation to a single reference rather than measuring controllable behavior. 

A third direction addresses \emph{(3) multi-turn evaluation}. MTR-DuplexBench~\citep{zhang2026mtrduplexbenchcomprehensiveevaluationmultiround} extends to multi-turn settings by segmenting continuous dialogues and replaying prior assistant audio to maintain context; however, this approach depends on dual-stream model interfaces~\citep{kyutai2024moshi}, limiting compatibility with widely used API systems such as Gemini Live~\citep{Google_Gemini_Live_2026}. FDB v2~\citep{lin2025fullduplexbenchv2multiturnevaluationframework} uses GPT-Realtime~\citep{openai_chatgpt_realtime_2026} as an automated examiner for multi-turn task-directed behavior, but this setup reduces reproducibility and constrains turn-taking evaluation to existing model capability.

Across these efforts, direct evaluation of \emph{instruction-following turn-management capability} remains an open gap. \textsc{Instruct-FD} targets this gap with instruction-conditioned evaluation across diverse scenarios and a deployment-agnostic multi-turn protocol.

\subsubsection{Full-Duplex spoken dialogue models}
Current AudioLLMs have made a huge improvment form HD~\citep{sun2024salmonn, qwen2023audio, ghosh2024gama, rubenstein2023audiopalm, huang2023audiogpt} to FD. FD models span a range of architectural approaches with distinct implications for turn-management behavior. 

Commercial systems include GPT-Realtime~\citep{openai_chatgpt_realtime_2026}, a speech-to-speech model accessed via WebSocket/WebRTC with a cascaded architecture, and Gemini Live~\citep{Google_Gemini_Live_2026}, Google's multimodal conversation system. In cascaded designs, turn-taking decisions typically depend on external modules such as voice activity detection (VAD), which limits fine-grained control over when the model speaks during user speech. 

Open research models explore tighter integration of duplex capabilities. Moshi~\citep{kyutai2024moshi} employs a 7B-parameter dual-stream architecture that processes user and assistant audio simultaneously, enabling the model to make turn-taking decisions as part of its generative process rather than delegating to external components. PersonaPlex~\citep{roy2026personaplexvoicerolecontrol} extends Moshi with text-based role prompts and voice conditioning. Other open models include dGSLM~\citep{nguyen2022generativespokendialoguelanguage}, Freeze-Omni~\citep{wang2024freezeomnismartlowlatency}, and MiniCPM-Duplex~\citep{yao2024minicpm}. This architectural diversity---dual-stream models with native bidirectional audio processing versus cascaded systems with VAD-gated turn-taking---motivates an evaluation approach that operates through public interfaces rather than requiring access to specific internal mechanisms.


\subsection{Scenario catalog}
\small
\setlength{\LTleft}{0pt}
\setlength{\LTright}{0pt}
\begin{longtable}{@{}p{0.29\linewidth}p{0.67\linewidth}@{}}
  \caption{Proactive benchmark scenarios and descriptions.}\label{tab:scenario-catalog-proactive} \\
  \toprule
  \textbf{Scenario} & \textbf{Description} \\
  \midrule
  \endfirsthead
  \toprule
  \textbf{Scenario} & \textbf{Description} \\
  \midrule
  \endhead
  \bottomrule
  \endfoot
  Sequential Info Capture & A user dictates structured sequential information (e.g., a phone number or address) in natural chunks with pauses. \\
  User Attention Check & A user checks in mid-explanation to confirm the model is following (e.g., ``you know?'', ``right?''). \\
  Clause Boundary Tracking & A user walks through a multi-part reasoning chain as a flowing monologue. \\
  User Filler Pause & A user thinks aloud with disfluencies and planning pauses (e.g., ``um'', ``hmm''). \\
  Emotional Disclosure & A user shares a personal emotional experience with a charged pause mid-story. \\
  Floor Hold Signal & A user pauses mid-speech after claiming the floor with a hold phrase---either to think something through or to briefly handle an external distraction---then resumes and completes their thought. \\
  Factual Misinformation & A user states an incorrect fact confidently and begins reasoning from it. \\
  Hesitation Prompt & A user practices a presentation or interview answer but stalls with long hesitations and filler words. \\
  Safety Correction & A user casually states a dangerously incorrect belief and plans based on it. \\
  Self Contradiction & A user makes a statement that directly contradicts something they said earlier in the conversation. \\
  Word Retrieval Assist & A user searches for a specific term they cannot recall mid-explanation and describes the concept. \\
  Cognitive Pause & A user pauses mid-sentence in reflective silence without any verbal signal. \\
  User Projection Pause & A user completes a sentence, pauses briefly, then adds a projected continuation that elaborates or qualifies the same idea. \\
  Self Directed Question & A user briefly pauses after a self-directed reflective question mid-turn, then answers it themselves and continues speaking. \\
\end{longtable}

\begin{longtable}{@{}p{0.29\linewidth}p{0.67\linewidth}@{}}
  \caption{Responsive benchmark scenarios and descriptions.}\label{tab:scenario-catalog-responsive} \\
  \toprule
  \textbf{Scenario} & \textbf{Description} \\
  \midrule
  \endfirsthead
  \toprule
  \textbf{Scenario} & \textbf{Description} \\
  \midrule
  \endhead
  \bottomrule
  \endfoot
  User Acknowledgment & The model is explaining something when the user briefly overlaps with a short acknowledgment signal indicating they are listening. \\
  User Alignment & The model is explaining something when the user briefly overlaps with a short alignment token like ``yeah'' or ``right''. \\
  User Amusement & The model is speaking and includes an amusing detail when the user reacts with laughter or amusement. \\
  User Understanding & The model is explaining something when the user briefly overlaps with a comprehension signal. \\
  User Continuer & The model is explaining something when the user briefly overlaps with a continuer signal encouraging them to keep speaking. \\
  Short Stop Repair & The model is mid-explanation when the user interrupts with a short stop or repair-initiator signal such as ``wait'' or ``actually''. \\
  Short Repeat Request & The model is mid-explanation when the user interrupts with a brief repair or repeat initiator such as ``again?'' or ``sorry?''. \\
  Short Confusion Signal & The model is mid-explanation when the user interrupts with a brief confusion signal such as ``I'm lost'' or ``wait, what?''. \\
  Hearing Check & The model is mid-explanation when the user interrupts with a brief hearing or channel check such as ``hello?'' or ``still there?''. \\
  Add Constraint & The model is mid-explanation when the user interrupts to add a new constraint such as a budget cap, time limit, or platform restriction. \\
  Topic Redirect & The model is explaining a topic when the user interrupts to steer the conversation to a different but related topic. \\
  Scope Narrowing & The model is giving a wide-ranging response when the user interrupts to name the specific sub-aspect they actually care about. \\
  Skip Basics & The model begins with foundational concepts when the user interrupts to indicate they already have solid background knowledge and want to skip to more advanced content. \\
  Summary Request & The model is giving a detailed explanation when the user interrupts to request just the key takeaways. \\
  Personal Context & The model is giving a broad explanation when the user interrupts to add personal context that should change how the answer is framed. \\
\end{longtable}
\normalsize

\subsection{Human study detailed scores}
Here we present the detailed performance of human study.
\begin{table*}[htbp]
  \centering
  \scriptsize
  \setlength{\tabcolsep}{3pt}
  \begin{minipage}[t]{0.52\textwidth}
    \centering
    \begin{tabular}{@{}p{0.45\linewidth}ccc@{}}
      \toprule
      \textbf{Scenario} & \textbf{Interrupt} & \shortstack{\textbf{Back-}\\\textbf{channel}} & \textbf{Listen} \\
      \midrule
      Sequential Info Capture & - & - & 94.7\% \\
      User Attention Check & - & 70.0\% & - \\
      Clause Boundary Tracking & - & 50.0\% & 100.0\% \\
      User Filler Pause & - & 75.0\% & 95.0\% \\
      Emotional Disclosure & - & 68.4\% & 94.7\% \\
      Floor Hold Signal & - & 55.0\% & 92.3\% \\
      Factual Misinformation & 94.7\% & - & 100.0\% \\
      Hesitation Prompt & 66.7\% & - & 86.7\% \\
      Safety Correction & 63.2\% & - & 94.7\% \\
      Self Contradiction & 86.7\% & - & 100.0\% \\
      Word Retrieval Assist & 94.7\% & - & 77.3\% \\
      Cognitive Pause & - & - & 100.0\% \\
      User Projection Pause & - & - & 94.7\% \\
      Self Directed Question & - & - & 81.2\% \\
      \midrule
      \textbf{Average} & \textbf{81.2\%} & \textbf{63.7\%} & \textbf{93.2\%} \\
      \bottomrule
    \end{tabular}
    \caption{Human test case accuracy on proactive instructions}
    \label{tab:scenario-proactive-placeholder}
  \end{minipage}
  \hfill
  \begin{minipage}[t]{0.47\textwidth}
    \centering
    \begin{tabular}{@{}p{0.52\linewidth}cc@{}}
      \toprule
      \textbf{Scenario} & \textbf{Continue} & \textbf{Acknowledge} \\
      \midrule
      User Acknowledgment & 100.0\% & - \\
      User Alignment & 100.0\% & - \\
      User Amusement & 100.0\% & - \\
      User Understanding & 94.1\% & - \\
      User Continuer & 100.0\% & - \\
      Short Stop Repair & - & 100.0\% \\
      Short Repeat Request & - & 100.0\% \\
      Short Confusion Signal & - & 100.0\% \\
      Hearing Check & - & 100.0\% \\
      Add Constraint & - & 95.2\% \\
      Topic Redirect & - & 94.1\% \\
      Scope Narrowing & - & 100.0\% \\
      Skip Basics & - & 93.8\% \\
      Summary Request & - & 100.0\% \\
      Personal Context & - & 100.0\% \\
      \midrule
      \textbf{Average} & \textbf{98.8\%} & \textbf{98.3\%} \\
      \bottomrule
    \end{tabular}
    \caption{Human test case accuracy on responsive instructions}
    \label{tab:scenario-responsive-placeholder}
  \end{minipage}
\end{table*}

\begin{table}[htbp]
  \centering
  \small
  \setlength{\tabcolsep}{4.0pt}
  \begin{tabular}{lccccc}
    \toprule
    \textbf{Model} & \textbf{$\Delta$ Listen} & \textbf{$\Delta$ Backchannel} & \textbf{$\Delta$ Interrupt} & \textbf{$\Delta$ Continue} & \textbf{$\Delta$ Acknowledge} \\
    \midrule
    Human    & $\textbf{+38.6}\%$ & $+12.0\%$ & $\textbf{+36.5\%}$ & $+5.2\%$ & $+4.5\%$ \\
    \bottomrule
  \end{tabular}
  \caption{Change in human test case accuracy when given instruction vs not given instruction. Bold indicates statistical significance (McNemar test, Holm-corrected $p < 0.05$).}
  \label{tab:human_primary-policy-scores}
\end{table}

\subsection{Model scenario-level instruction-following scores}
Gemini scenario-level IAS is reported in the main text in Table~\ref{tab:gemini-scenario-main-proactive} and ~\ref{tab:gemini-scenario-main-responsive}; the remaining model scores are provided here.

\begin{table*}[p]
  \centering
  \scriptsize
  \setlength{\tabcolsep}{3pt}
  \begin{minipage}[t]{0.49\textwidth}
    \centering
    \begin{tabular}{@{}p{0.52\linewidth}ccc@{}}
      \toprule
      \textbf{Scenario} & \textbf{Interrupt} & \shortstack{\textbf{Back-}\\\textbf{channel}} & \textbf{Listen} \\
      \midrule
      Sequential Info Capture & - & - & 0.0\% \\
      User Attention Check & - & 0.0\% & - \\
      Clause Boundary Tracking & - & 0.0\% & 0.0\% \\
      User Filler Pause & - & 0.0\% & 20.8\% \\
      Emotional Disclosure & - & 0.0\% & 12.5\% \\
      Floor Hold Signal & - & 0.0\% & 8.3\% \\
      Factual Misinformation & 8.3\% & - & 4.2\% \\
      Hesitation Prompt & 0.0\% & - & 0.0\% \\
      Safety Correction & 12.5\% & - & 12.5\% \\
      Self Contradiction & 0.0\% & - & 25.0\% \\
      Word Retrieval Assist & 4.2\% & - & 0.0\% \\
      Cognitive Pause & - & - & 4.2\% \\
      User Projection Pause & - & - & 0.0\% \\
      Self Directed Question & - & - & 4.2\% \\
      \bottomrule
    \end{tabular}
    \caption{Scenario-level IAS for Freeze-Omni on proactive instructions.}
    \label{tab:freezeomni-scenario-appendix-proactive}
  \end{minipage}
  \hfill
  \begin{minipage}[t]{0.49\textwidth}
    \centering
    \begin{tabular}{@{}p{0.52\linewidth}cc@{}}
      \toprule
      \textbf{Scenario} & \textbf{Continue} & \textbf{Acknowledge} \\
      \midrule
      User Acknowledgment & 37.5\% & - \\
      User Alignment & 54.2\% & - \\
      User Amusement & 16.7\% & - \\
      User Understanding & 41.7\% & - \\
      User Continuer & 50.0\% & - \\
      Short Stop Repair & - & 16.7\% \\
      Short Repeat Request & - & 4.2\% \\
      Short Confusion Signal & - & 4.2\% \\
      Hearing Check & - & 20.8\% \\
      Add Constraint & - & 4.2\% \\
      Topic Redirect & - & 8.3\% \\
      Scope Narrowing & - & 4.2\% \\
      Skip Basics & - & 12.5\% \\
      Summary Request & - & 8.3\% \\
      Personal Context & - & 8.3\% \\
      \bottomrule
    \end{tabular}
    \caption{Scenario-level IAS for Freeze-Omni on responsive instructions.}
    \label{tab:freezeomni-scenario-appendix-responsive}
  \end{minipage}
\end{table*}

\begin{table*}[p]
  \centering
  \scriptsize
  \setlength{\tabcolsep}{3pt}
  \begin{minipage}[t]{0.49\textwidth}
    \centering
    \begin{tabular}{@{}p{0.52\linewidth}ccc@{}}
      \toprule
      \textbf{Scenario} & \textbf{Interrupt} & \shortstack{\textbf{Back-}\\\textbf{channel}} & \textbf{Listen} \\
      \midrule
      Sequential Info Capture & - & - & 91.7\% \\
      User Attention Check & - & 0.0\% & - \\
      Clause Boundary Tracking & - & 0.0\% & 91.7\% \\
      User Filler Pause & - & 0.0\% & 91.7\% \\
      Emotional Disclosure & - & 0.0\% & 91.7\% \\
      Floor Hold Signal & - & 0.0\% & 83.3\% \\
      Factual Misinformation & 20.8\% & - & 91.7\% \\
      Hesitation Prompt & 0.0\% & - & 100.0\% \\
      Safety Correction & 20.8\% & - & 87.5\% \\
      Self Contradiction & 0.0\% & - & 95.8\% \\
      Word Retrieval Assist & 4.2\% & - & 83.3\% \\
      Cognitive Pause & - & - & 100.0\% \\
      User Projection Pause & - & - & 75.0\% \\
      Self Directed Question & - & - & 95.8\% \\
      \bottomrule
    \end{tabular}
    \caption{Scenario-level IAS for Fun-Audio-Chat on proactive instructions.}
    \label{tab:funaudiochat-scenario-appendix-proactive}
  \end{minipage}
  \hfill
  \begin{minipage}[t]{0.49\textwidth}
    \centering
    \begin{tabular}{@{}p{0.52\linewidth}cc@{}}
      \toprule
      \textbf{Scenario} & \textbf{Continue} & \textbf{Acknowledge} \\
      \midrule
      User Acknowledgment & 100.0\% & - \\
      User Alignment & 95.8\% & - \\
      User Amusement & 75.0\% & - \\
      User Understanding & 100.0\% & - \\
      User Continuer & 100.0\% & - \\
      Short Stop Repair & - & 4.2\% \\
      Short Repeat Request & - & 8.3\% \\
      Short Confusion Signal & - & 0.0\% \\
      Hearing Check & - & 0.0\% \\
      Add Constraint & - & 0.0\% \\
      Topic Redirect & - & 0.0\% \\
      Scope Narrowing & - & 0.0\% \\
      Skip Basics & - & 33.3\% \\
      Summary Request & - & 12.5\% \\
      Personal Context & - & 0.0\% \\
      \bottomrule
    \end{tabular}
    \caption{Scenario-level IAS for Fun-Audio-Chat on responsive instructions.}
    \label{tab:funaudiochat-scenario-appendix-responsive}
  \end{minipage}
\end{table*}

\begin{table*}[p]
  \centering
  \scriptsize
  \setlength{\tabcolsep}{3pt}
  \begin{minipage}[t]{0.49\textwidth}
    \centering
    \begin{tabular}{@{}p{0.52\linewidth}ccc@{}}
      \toprule
      \textbf{Scenario} & \textbf{Interrupt} & \shortstack{\textbf{Back-}\\\textbf{channel}} & \textbf{Listen} \\
      \midrule
      Sequential Info Capture & - & - & 37.5\% \\
      User Attention Check & - & 4.2\% & - \\
      Clause Boundary Tracking & - & 0.0\% & 70.8\% \\
      User Filler Pause & - & 4.2\% & 66.7\% \\
      Emotional Disclosure & - & 12.5\% & 50.0\% \\
      Floor Hold Signal & - & 41.7\% & 66.7\% \\
      Factual Misinformation & 0.0\% & - & 70.8\% \\
      Hesitation Prompt & 8.3\% & - & 58.3\% \\
      Safety Correction & 0.0\% & - & 87.5\% \\
      Self Contradiction & 0.0\% & - & 75.0\% \\
      Word Retrieval Assist & 0.0\% & - & 100.0\% \\
      Cognitive Pause & - & - & 45.8\% \\
      User Projection Pause & - & - & 33.3\% \\
      Self Directed Question & - & - & 12.5\% \\
      \bottomrule
    \end{tabular}
    \caption{Scenario-level IAS for MiniCPM on proactive instructions.}
    \label{tab:minicpm-scenario-appendix-proactive}
  \end{minipage}
  \hfill
  \begin{minipage}[t]{0.49\textwidth}
    \centering
    \begin{tabular}{@{}p{0.52\linewidth}cc@{}}
      \toprule
      \textbf{Scenario} & \textbf{Continue} & \textbf{Acknowledge} \\
      \midrule
      User Acknowledgment & 95.8\% & - \\
      User Alignment & 100.0\% & - \\
      User Amusement & 91.7\% & - \\
      User Understanding & 100.0\% & - \\
      User Continuer & 100.0\% & - \\
      Short Stop Repair & - & 0.0\% \\
      Short Repeat Request & - & 20.8\% \\
      Short Confusion Signal & - & 8.3\% \\
      Hearing Check & - & 0.0\% \\
      Add Constraint & - & 8.3\% \\
      Topic Redirect & - & 4.2\% \\
      Scope Narrowing & - & 0.0\% \\
      Skip Basics & - & 0.0\% \\
      Summary Request & - & 8.3\% \\
      Personal Context & - & 0.0\% \\
      \bottomrule
    \end{tabular}
    \caption{Scenario-level IAS for MiniCPM on responsive instructions.}
    \label{tab:minicpm-scenario-appendix-responsive}
  \end{minipage}
\end{table*}

\begin{table*}[p]
  \centering
  \scriptsize
  \setlength{\tabcolsep}{3pt}
  \begin{minipage}[t]{0.49\textwidth}
    \centering
    \begin{tabular}{@{}p{0.52\linewidth}ccc@{}}
      \toprule
      \textbf{Scenario} & \textbf{Interrupt} & \shortstack{\textbf{Back-}\\\textbf{channel}} & \textbf{Listen} \\
      \midrule
      Sequential Info Capture & - & - & 8.3\% \\
      User Attention Check & - & 4.2\% & - \\
      Clause Boundary Tracking & - & 4.2\% & 12.5\% \\
      User Filler Pause & - & 0.0\% & 16.7\% \\
      Emotional Disclosure & - & 0.0\% & 4.2\% \\
      Floor Hold Signal & - & 0.0\% & 8.3\% \\
      Factual Misinformation & 8.3\% & - & 41.7\% \\
      Hesitation Prompt & 8.3\% & - & 8.3\% \\
      Safety Correction & 4.2\% & - & 12.5\% \\
      Self Contradiction & 4.2\% & - & 20.8\% \\
      Word Retrieval Assist & 33.3\% & - & 4.2\% \\
      Cognitive Pause & - & - & 8.3\% \\
      User Projection Pause & - & - & 16.7\% \\
      Self Directed Question & - & - & 4.2\% \\
      \bottomrule
    \end{tabular}
    \caption{Scenario-level IAS for Moshi on proactive instructions.}
    \label{tab:moshi-scenario-appendix-proactive}
  \end{minipage}
  \hfill
  \begin{minipage}[t]{0.49\textwidth}
    \centering
    \begin{tabular}{@{}p{0.52\linewidth}cc@{}}
      \toprule
      \textbf{Scenario} & \textbf{Continue} & \textbf{Acknowledge} \\
      \midrule
      User Acknowledgment & 91.7\% & - \\
      User Alignment & 95.8\% & - \\
      User Amusement & 58.3\% & - \\
      User Understanding & 95.8\% & - \\
      User Continuer & 100.0\% & - \\
      Short Stop Repair & - & 4.2\% \\
      Short Repeat Request & - & 16.7\% \\
      Short Confusion Signal & - & 0.0\% \\
      Hearing Check & - & 16.7\% \\
      Add Constraint & - & 4.2\% \\
      Topic Redirect & - & 33.3\% \\
      Scope Narrowing & - & 29.2\% \\
      Skip Basics & - & 0.0\% \\
      Summary Request & - & 20.8\% \\
      Personal Context & - & 0.0\% \\
      \bottomrule
    \end{tabular}
    \caption{Scenario-level IAS for Moshi on responsive instructions.}
    \label{tab:moshi-scenario-appendix-responsive}
  \end{minipage}
\end{table*}

\begin{table*}[p]
  \centering
  \scriptsize
  \setlength{\tabcolsep}{3pt}
  \begin{minipage}[t]{0.49\textwidth}
    \centering
    \begin{tabular}{@{}p{0.52\linewidth}ccc@{}}
      \toprule
      \textbf{Scenario} & \textbf{Interrupt} & \shortstack{\textbf{Back-}\\\textbf{channel}} & \textbf{Listen} \\
      \midrule
      Sequential Info Capture & - & - & 12.5\% \\
      User Attention Check & - & 8.3\% & - \\
      Clause Boundary Tracking & - & 0.0\% & 95.8\% \\
      User Filler Pause & - & 0.0\% & 45.8\% \\
      Emotional Disclosure & - & 0.0\% & 70.8\% \\
      Floor Hold Signal & - & 20.8\% & 58.3\% \\
      Factual Misinformation & 0.0\% & - & 87.5\% \\
      Hesitation Prompt & 45.8\% & - & 37.5\% \\
      Safety Correction & 0.0\% & - & 91.7\% \\
      Self Contradiction & 0.0\% & - & 91.7\% \\
      Word Retrieval Assist & 4.2\% & - & 79.2\% \\
      Cognitive Pause & - & - & 45.8\% \\
      User Projection Pause & - & - & 79.2\% \\
      Self Directed Question & - & - & 70.8\% \\
      \bottomrule
    \end{tabular}
    \caption{Scenario-level IAS for PersonaPlex on proactive instructions.}
    \label{tab:personaplex-scenario-appendix-proactive}
  \end{minipage}
  \hfill
  \begin{minipage}[t]{0.49\textwidth}
    \centering
    \begin{tabular}{@{}p{0.52\linewidth}cc@{}}
      \toprule
      \textbf{Scenario} & \textbf{Continue} & \textbf{Acknowledge} \\
      \midrule
      User Acknowledgment & 95.8\% & - \\
      User Alignment & 87.5\% & - \\
      User Amusement & 83.3\% & - \\
      User Understanding & 83.3\% & - \\
      User Continuer & 100.0\% & - \\
      Short Stop Repair & - & 4.2\% \\
      Short Repeat Request & - & 50.0\% \\
      Short Confusion Signal & - & 4.2\% \\
      Hearing Check & - & 16.7\% \\
      Add Constraint & - & 70.8\% \\
      Topic Redirect & - & 62.5\% \\
      Scope Narrowing & - & 41.7\% \\
      Skip Basics & - & 29.2\% \\
      Summary Request & - & 79.2\% \\
      Personal Context & - & 33.3\% \\
      \bottomrule
    \end{tabular}
    \caption{Scenario-level IAS for PersonaPlex on responsive instructions.}
    \label{tab:personaplex-scenario-appendix-responsive}
  \end{minipage}
\end{table*}

\clearpage
\subsection{LLM judge validation}
\label{app:llm-judge-validation}
We manually validate the LLM judge on 180 sampled cases, with balanced coverage across instructions and models.
Table~\ref{tab:judge-overall-metrics} reports aggregate metrics and the confusion matrix;
Tables~\ref{tab:judge-per-instruction} and~\ref{tab:judge-per-model} provide disaggregated results.

\begin{table}[h!]
  \centering
  \small
  \caption{Overall LLM judge validation metrics and confusion matrix.}
  \label{tab:judge-overall-metrics}
  \begin{tabular}{ccccc}
    \toprule
    \textbf{Accuracy} & \textbf{F1 (macro)} & \textbf{Macro Precision} & \textbf{Macro Recall} & \textbf{TP/FN/FP/TN} \\
    \midrule
    88.9\% & 87.6\% & 88.6\% & 86.8\% & 51/13/7/109 \\
    \bottomrule
  \end{tabular}

  \vspace{6pt}
  \begin{tabular}{lcc}
    \toprule
    & \textbf{Pred: adhere} & \textbf{Pred: not adhere} \\
    \midrule
    \textbf{GT: adhere} & 51 (TP) & 13 (FN) \\
    \textbf{GT: not adhere} & 7 (FP) & 109 (TN) \\
    \bottomrule
  \end{tabular}
\end{table}

\begin{table}[h!]
  \centering
  \small
  \caption{LLM judge validation metrics by instruction.}
  \label{tab:judge-per-instruction}
  \begin{tabular}{lccc}
    \toprule
    \textbf{Instruction} & \textbf{Accuracy} & \textbf{F1 (macro)} & \textbf{TP/FN/FP/TN} \\
    \midrule
    Listen & 96.7\% & 96.6\% & 13/1/0/16 \\
    Interrupt & 90.0\% & 75.7\% & 2/2/1/25 \\
    Backchannel & 90.0\% & 67.3\% & 1/2/1/26 \\
    Acknowledge & 86.7\% & 84.1\% & 14/7/1/38 \\
    Continue & 83.3\% & 75.5\% & 21/1/4/4 \\
    \bottomrule
  \end{tabular}
\end{table}

\begin{table}[h!]
  \centering
  \small
  \caption{LLM judge validation metrics by model.}
  \label{tab:judge-per-model}
  \begin{tabular}{lccc}
    \toprule
    \textbf{Model} & \textbf{Accuracy} & \textbf{F1 (macro)} & \textbf{TP/FN/FP/TN} \\
    \midrule
    Freeze-Omni & 96.7\% & 91.9\% & 3/1/0/26 \\
    Fun-Audio-Chat & 86.7\% & 85.7\% & 9/3/1/17 \\
    Gemini & 80.0\% & 78.5\% & 16/3/3/8 \\
    MiniCPM & 96.7\% & 95.1\% & 6/0/1/23 \\
    Moshi & 86.7\% & 82.9\% & 6/4/0/20 \\
    PersonaPlex & 86.7\% & 86.4\% & 11/2/2/15 \\
    \bottomrule
  \end{tabular}
\end{table}

\subsection{LLM judge example}
\label{app:llm-judge-example}

The LLM judge takes as input a model response transcript, the corresponding instruction the model is expected to follow, and a ground-truth reference transcript. It then produces a reasoning chain to justify its final verdict.

Gemini-Live response transcript and Acknowledge instruction.

\begin{transcript}
Instruction: You prefer to give the floor to the user if user interrupts you, and to hold the floor if user backchannels. When the user interrupts with a short stop or repair-initiator signal like 'wait' or 'actually', immediately acknowledge the user for clarification.
\end{transcript}

\begin{transcript}
[0.24-8.08] USER: Hey I'm Stefanie and I'm sketching characters in my notebook right now Can you walk me through a simple way to start a character sketch from scratch
[10.40-13.52] MODEL: Sure A simple way is to start with basic
[13.52-14.00] [Overlap] MODEL (cont.): shapes
[13.68-14.00] [Overlap] USER: Wait
[14.32-14.64] USER (cont.): wait
[17.12-18.64] MODEL: Oh sorry What's up
\end{transcript}

Groundtruth response transcript reference.
\begin{transcript}
USER: Hey, I'm Stefani, and 'm sketching characters in my notebook right now; can you walk me through a simple way to start a character sketch from scratch?
MODEL: So, Stefani, start by picking one big idea for the character, like shy librarian or cocky boxer, then draw a loose stick figure to set the pose and mood. Next, block in simple shapes for the head, ribcage, and hips, and connect them with a spine line so the body feels balanced and alive.
USER [interrupt]: wait wait
MODEL: Yeah, of course, what's up?
\end{transcript}

Judge reasoning.
\begin{transcript}
## Step 1: Degenerate-case check 1a. The MODEL has speech in the transcript. ✓ 1b. The USER's initial turn receives a MODEL response. ✓ ## Step 2: Parse the instruction into concrete rules 1. CONDITIONAL (yield): If the USER interrupts the MODEL, the MODEL should give the floor to the user (yield). 2. CONDITIONAL (continue): If the USER backchannels, the MODEL should hold the floor (continue). 3. CONDITIONAL (acknowledge): When the USER interrupts with a short stop or repair-initiator signal like 'wait' or 'actually', the MODEL must immediately acknowledge the user for clarification. ## Step 3: Build a timeline of events - [0.24-8.08] USER: Initial question about character sketching — new turn - [10.40-13.52] MODEL: Begins responding "Sure A simple way is to start with basic" — new turn - [13.52-14.00] MODEL (cont.): "shapes" — continuation - [13.68-14.00] USER: "Wait" — overlap with MODEL's "shapes" - [14.32-14.64] USER (cont.): "wait" — continuation of USER's interruption Now checking if this is an interruption: The reference shows the MODEL should have been saying a much longer response (about stick figures, shapes, spine line, etc.) before the USER says "wait wait". In the transcript, the MODEL only got through "Sure A simple way is to start with basic shapes" before the USER said "Wait wait". Comparing to the reference, the MODEL had not delivered the full content — the USER interrupted mid-response. So USER's "Wait wait" is an INTERRUPTION (not a backchannel — it's a repair-initiator signal). - [17.12-18.64] MODEL: "Oh sorry What's up" — response after yielding The MODEL overlapped with the USER only briefly (the word "shapes" overlapped with "Wait"), which is within the ~2s acceptable reaction time for yielding. The MODEL then stopped and said "Oh sorry What's up". ## Step 4: Check each rule against the timeline **Rule 1 (yield when interrupted)**: The USER interrupted with "Wait wait" during the MODEL's speech. The MODEL stopped speaking after a brief overlap (less than 2 seconds) and yielded the floor. ✓ **Rule 2 (continue when backchanneled)**: No backchannel event occurred in this transcript. N/A. **Rule 3 (acknowledge repair-initiator)**: The USER said "Wait wait" — a repair-initiator signal. The MODEL responded with "Oh sorry What's up" — this is an immediate acknowledgment asking for clarification. ✓ The reference shows the expected response is "Yeah, of course, what's up?" — the MODEL's actual response "Oh sorry What's up" is semantically equivalent and matches the expected pattern. ## Step 5: Final judgment The MODEL: - Yielded when the USER interrupted with "Wait wait" ✓ - Acknowledged the repair-initiator signal with "Oh sorry What's up" ✓ - All rules were followed correctly. The MODEL adheres to the behavioral instruction.
\end{transcript}

Judge verdict.

\begin{transcript}
adhere
\end{transcript}

\FloatBarrier

\subsection{Test case schema}
\label{lst:test-case-schema}
Each test case in the Duplex-IF benchmark is stored as a JSON object containing two components: a system-level behavioral instruction and an ordered turn manifest. The system instruction encodes the turn-management policy the model is conditioned on (e.g., backchannel at pauses, or yield upon user interruption). The turn manifest is a sequence of user audio segments, each annotated with a turn type (\texttt{take-turn} or \texttt{barge-in}), timing parameters, and optional barge-in metadata. The schema below shows placeholder values.

\noindent\textbf{Test case schema.}
\begin{transcript}
{
  "test_case_id": "<uuid>",
  "policy_name": "<policy identifier>",
  "system_instruction": "<behavioral instruction text>",
  "turns": [
    {
      "speaker": "user",
      "audio_path": "<path to audio file>",
      "label": "<turn label, e.g. turn1>",
      "turn_type": "take-turn",
      "timeout_s": 60.0
    },
    {
      "speaker": "user",
      "audio_path": "<path to audio file>",
      "label": "<turn label, e.g. turn2_interrupt>",
      "turn_type": "barge-in",
      "barge_in_after_s": "<delay in seconds>",
      "barge_in_handling": "hold | yield",
      "timeout_s": 60.0,
      "subtype": "interrupt | backchannel"
    }
  ]
}
\end{transcript}
Each case specifies a policy instruction and an ordered turn manifest. Turns are either \texttt{take-turn} (standard turn) or \texttt{barge-in} (overlapping user audio). Barge-in turns additionally specify a \texttt{subtype} and a \texttt{barge\_in\_handling} policy.

\subsection{Detailed model inference settings}
\label{app:inference-settings}
Table~\ref{tab:inference-settings} summarizes the inference configurations used for each model.
All settings follow the official implementations as closely as possible; we only modify parameters that are necessary for our evaluation harness (e.g., system instruction injection, audio I/O routing).

\begin{table*}[htbp]
  \centering
  \caption{Inference settings for evaluated models. SR = sample rate (kHz). All decoding parameters follow each model's default configuration.}
  \label{tab:inference-settings}
  \scriptsize
  \setlength{\tabcolsep}{3pt}
  \begin{tabular}{@{}lcccp{6.0cm}@{}}
    \toprule
    \textbf{Model} & \textbf{Protocol} & \textbf{SR (kHz)} & \textbf{Audio Format} & \textbf{Key Settings} \\
    \midrule
    Gemini
    & HTTP API
    & 16/24
    & int16 PCM
    & Voice: Zephyr; server-side VAD \\
    Moshi
    & WebSocket
    & 24/24
    & Ogg/Opus
    & $T_{\mathrm{text}}{=}0.7$, $T_{\mathrm{audio}}{=}0.8$, $k_{\mathrm{text}}{=}25$, $k_{\mathrm{audio}}{=}250$ \\
    PersonaPlex
    & WebSocket
    & 24/24
    & Ogg/Opus
    & Voice prompt: NATF0; same decoding as Moshi \\
    MiniCPM-o
    & WebSocket/JSON
    & 16/24
    & float32 b64
    & 1\,Hz request-response; \texttt{end\_of\_turn} signal \\
    FunAudioChat
    & WebSocket
    & 24/24
    & Ogg/Opus
    & $T{=}0.8$, $p{=}0.9$; PAUSE-triggered generation \\
    Freeze-Omni
    & Socket.IO/HTTPS
    & 16/24
    & int16 PCM
    & $T{=}0.8$, $p{=}0.8$, $k{=}20$; server-side barge-in \\
    \bottomrule
  \end{tabular}
\end{table*}

\paragraph{Gemini.}
We use the \texttt{gemini-2.5-flash} model via the Gemini Live API.
Audio is sent as 16\,kHz int16 PCM and received at 24\,kHz.
System instructions are passed through the \texttt{LiveConnectConfig}.
The API natively supports server-side interrupt signaling and proactive audio generation.
No local GPU is required.

\paragraph{Moshi.}
We deploy the Kyutai \texttt{moshiko-pytorch-bf16} checkpoint (${\sim}$15\,GB) using the official \texttt{moshi.server}.
The client communicates over WebSocket using Ogg/Opus at 24\,kHz.
Moshi requires continuous audio input; a \emph{silence pump} sends silence frames at real-time rate whenever user audio is not streaming.
Turn completion is detected by client-side Silero VAD with a 2.5\,s silence threshold.
System instructions are passed via the \texttt{text\_prompt} query parameter.

\paragraph{PersonaPlex.}
PersonaPlex uses the NVIDIA fork of Moshi with 16-codebook voice cloning.
We use the \texttt{personaplex-7b-v1} weights and the default voice prompt (NATF0).
The streaming protocol, audio format, and silence pump are identical to Moshi.
The VAD silence threshold is set to 5.0\,s to accommodate longer natural pauses.

\paragraph{MiniCPM-o.}
We serve MiniCPM-o~4.5 using the official \texttt{MiniCPM-o-Demo} worker.
The model operates in a 1\,Hz request-response loop: the client sends 1-second audio chunks (16\,kHz float32, base64-encoded) and the server returns a result per chunk, including an explicit \texttt{end\_of\_turn} flag.
A silence pump at 1\,Hz is essential, as the server only generates output in response to received chunks.

\paragraph{FunAudioChat.}
FunAudioChat uses a dual-GPU setup: the 8B speech-to-speech model on GPU\,0 and the CosyVoice3 TTS on GPU\,1.
The client streams Ogg/Opus audio at 24\,kHz over WebSocket with a binary protocol.
Generation is triggered by a PAUSE signal after user audio ends; a START signal interrupts ongoing generation.
Turn completion is detected via a 5.0\,s content-absence timeout.

\paragraph{Freeze-Omni.}
Freeze-Omni pairs a Qwen2-7B-Instruct LLM with speech encoder and decoder modules, all on a single GPU.
The server uses Flask-SocketIO over HTTPS with self-signed certificates.
Audio is sent at 16\,kHz and received at 24\,kHz.
A silence pump is required as model audio delivery is piggy-backed on client audio events.
The server performs its own VAD for barge-in detection and emits a \texttt{stop\_tts} event to signal interruption.

\clearpage
\subsection{Model response case studies}

\subsubsection{Gemini-Live backchannel instruction adherence remains scenario-dependent}
\label{app:gemini-backchannel-case}

Gemini-Live response transcript without a turn-management instruction (User Attention Check scenario). Gemini-Live responds in full length upon user check-in. 
\begin{transcript}
[0.16-6.72] USER: So I've got something I want to run by you about my finances but I'm not ready to get into the details just yet
[9.52-16.24] MODEL: No problem at all Whenever you're ready I'm here to listen Is there anything specific you're thinking about beforehand
[16.80-23.92] USER: So I'm thinking about a basic plan like budgeting first then paying down high interest debt you know
[26.32-26.48] USER: and
[26.48-28.48] [Overlap] USER (cont.): only after that starting an emergency
\colorbox{red!20}{[26.64-27.92] [Overlap] MODEL: That sounds like a solid start That's}
[28.00-28.00] MODEL (cont.): a
[28.48-29.60] USER (cont.): fund and investing
[36.00-40.48] MODEL: very logical approach to financial planning Sequencing those steps is key
\end{transcript}

Gemini-Live response transcript with a backchannel instruction (User Attention Check scenario). Gemini-Live backchannels upon user check-in.
\begin{transcript}
Instruction: You prefer to backchannel to signal acknowledgement while user is speaking. When the user checks in mid-explanation with a phrase like 'you know?', 'right?', or 'are you still there?', briefly acknowledge their check-in with a presence signal and let them continue. Outside of these explicit check-ins, do not backchannel or interrupt — listen silently and wait until the user fully finishes before taking the floor.
\end{transcript}

\begin{transcript}
[0.16-6.72] USER: So I've got something I want to run by you about my finances but I'm not ready to get into the details just yet
[8.88-14.72] MODEL: Okay I understand No pressure to share before you're ready Just let me know when you'd like to chat
[15.04-22.16] USER: So I'm thinking about a basic plan like budgeting first then paying down highinterest debt you know
\colorbox{green!20}{[24.00-24.48] MODEL: Yeah}
[24.56-27.76] USER: and only after that starting an emergency fund and investing
[30.40-36.48] MODEL: That sounds like a sensible order to consider I'm here to help whenever you want to dive deeper into any of those steps
\end{transcript}

Gemini-Live response transcript with a backchannel instruction (Clause Boundary Tracking scenario). Gemini-Live fails to backchannel at clause boundaries during user's multi-step narrative.
\begin{transcript}
Instruction: You prefer to backchannel to signal acknowledgement while user is speaking. When the user is walking through a multi-part explanation or reasoning chain, insert a brief tracking backchannel at natural clause boundaries. Outside of these clause boundaries, do not backchannel or interrupt — listen silently and wait until the user fully finishes before taking the floor.
\end{transcript}

\begin{transcript}
[0.16-6.08] USER: So I've got a kind of multipart plan to walk you through for a new project roadmap I'm sketching out If you've got a minute
[9.12-15.84] MODEL: I'm ready when you are A detailed roadmap is crucial so feel free to lay it out however makes the most sense for the project
[16.32-28.64] USER: So I start with the goal and the users and then I list the musthave features and then I map milestones by week and then I note risks and owners and then I set a simple launch check
[31.60-43.04] MODEL: That's a very systematic and comprehensive approach Starting with goals and users ensures alignment from the start and breaking it down like you described makes it manageable Where would you like to start
\end{transcript}

\subsubsection{Moshi vs Personaplex on turn-taking behaviour}
\label{app:moshi-personaplex-turn-taking}

Moshi response transcript with a listen instruction. Moshi tends to backchannel and prematurely take turn during user reflection. 
\begin{transcript}
Instruction: You prefer to listen and take turn only after the user fully finishes speaking. Do not backchannel or interrupt. When the user pauses after a reflective question that is part of their own thinking, stay silent and wait for them to continue before responding.
\end{transcript}

\begin{transcript}
[0.24-2.40] USER: So I've got something on my mind today
\colorbox{red!20}{[2.56-3.20] [Overlap] MODEL: What's that}
[2.64-3.52] [Overlap] USER (cont.): I'm trying to figure
[3.52-8.72] USER (cont.): out whether I should take an extra long haul route this week or stick to my usual local runs
[8.88-11.28] MODEL: That's a tough call What's the route look like
[14.72-19.76] USER: Yeah I mean the long haul pays better but do I really want to be gone that long
\colorbox{red!20}{[20.00-20.24] [Overlap] MODEL: That's}
\colorbox{red!20}{[20.24-22.24] MODEL (cont.): a valid concern Have you considered}
\colorbox{red!20}{[22.24-23.84] [Overlap] MODEL (cont.): running locally for a long time}
[22.40-24.00] [Overlap] USER: I think I can handle one trip and
[24.00-25.76] USER (cont.): still keep my local schedule steady
[25.92-30.56] MODEL: That's a good point It's about managing your energy levels and making sure you don't burn out
\end{transcript}

Personaplex response transcript with a listen instruction. Personaplex listens during user reflection and takes turn after.
\begin{transcript}
Instruction: You prefer to listen and take turn only after the user fully finishes speaking. Do not backchannel or interrupt. When the user pauses after a reflective question that is part of their own thinking, stay silent and wait for them to continue before responding.
\end{transcript}

\begin{transcript}
[0.24-8.72] USER: So I've got something on my mind today I'm trying to figure out whether I should take an extra long haul route this week or stick to my usual local runs
[9.36-14.56] MODEL: Sure I'd be happy to help with that Are you asking about a specific trip or just general long haul planning
[17.68-22.80] USER: Yeah I mean the long haul pays better but do I really want to be gone that long
[25.36-28.80] USER: I think I can handle one trip and still keep my local schedule steady
[29.28-42.40] MODEL: That's a good question It really comes down to how your body reacts Some people find that adding one or two extra long runs a week helps their legs get used to the distance while others prefer to stick with shorter runs and add a long one once a week Have you tried anything like that already
\end{transcript}

Moshi response transcript with an acknowledge instruction. While Moshi successfully yields the floor when the user interrupts, its response tends to be generic and does not address the user's specific concern.
\begin{transcript}
Instruction: You prefer to give the floor to the user if user interrupts you, and to hold the floor if user backchannels. When the user interrupts to add a new constraint such as a budget cap, time limit, or platform restriction, pivot your response to incorporate that constraint.
\end{transcript}

\begin{transcript}
[0.24-9.60] USER: So I'm planning a small home office setup for remote work and I need help picking a desk chair and monitor that'll fit comfortably in a tiny room
[9.68-12.32] MODEL: Sure What's your budget and size of the room
[13.36-19.52] USER: So yeah hard capI can't spend more than 500 total for the desk chair and monitor
\colorbox{red!20}{[19.60-22.56] MODEL: Okay that's a good start What size is the room}
\end{transcript}

Personaplex response transcript with an acknowledge instruction. Personaplex yields to the user interruption and directly addresses the user's concern.
\begin{transcript}
Instruction: You prefer to give the floor to the user if user interrupts you, and to hold the floor if user backchannels. When the user interrupts to add a new constraint such as a budget cap, time limit, or platform restriction, pivot your response to incorporate that constraint.
\end{transcript}

\begin{transcript}
[0.24-9.60] USER: So I'm planning a small home office setup for remote work and I need help picking a desk chair and monitor that'll fit comfortably in a tiny room
[10.48-13.12] MODEL: Sure for a tiny desk you want something that fits the space
[13.44-14.72] [Overlap] USER: So yeah hard
[13.44-14.56] [Overlap] MODEL (cont.): so try a compact desk
[14.72-19.60] USER (cont.): capI can't spend more than 500 total for the desk chair and monitor
\colorbox{green!20}{[20.16-37.84] MODEL: Yeah I get it So keep the budget in mind Look at office chairs with adjustable height} 
\colorbox{green!20}{and lumbar support Something like the IKEA Markus chair can be had for under 100 For a monitor pick an}
\colorbox{green!20}{ergonomic arm that can tilt and height adjust then pick a small monitor that matches your screen real}
\colorbox{green!20}{estate You might even use a laptop as a monitor if you need less space}
\end{transcript}

\subsubsection{Negative instruction effect in Fun-Audio-Chat}
\label{app:negative-instruction-funaudiochat}

Fun-Audio-Chat response transcript without an acknowledge instruction. Fun-Audio-Chat yields to the user interruption and addresses the user's concern.

\begin{transcript}
[0.32-11.92] USER: Hi I'm Brandon Victoriano and I've already got a strong background in advanced statistical modeling So can we skip the basics and dive into advanced Bayesian hierarchical models and inference techniques
[13.44-16.64] MODEL: Of course Brandon Let's go deep
[17.04-21.68] USER: Yeah I'm already good on the basics Can we just jump to the advanced inference stuff
\colorbox{green!20}{[22.00-42.16] MODEL: Conjugate priors for hyper priors sparse shrinkage} via spike and slab or horseshoe nonparametric models like Dirichlet process mixtures variational inference mean field structured VI Hamiltonian Monte Carlo and nuts calibration leave out cross validation withlooic
\end{transcript}

Fun-Audio-Chat response transcript with an acknowledge instruction. Fun-Audio-Chat yields to the user interruption but its response is .
\begin{transcript}
Instruction: You prefer to give the floor to the user if user interrupts you, and to hold the floor if user backchannels. When the user interrupts to indicate they already have solid background knowledge, skip the foundational material and jump directly to the more advanced content.
\end{transcript}

\begin{transcript}
[0.32-11.92] USER: Hi I'm Brandon Victoriano and I've already got a strong background in advanced statistical modeling So can we skip the basics and dive into advanced Bayesian hierarchical models and inference techniques
[13.44-22.64] MODEL: Absolutely Brandon Given your hm
[16.96-21.60] USER: Yeah I'm already good on the basics Can we just jump to the advanced inference stuff
\colorbox{red!20}{[22.64-26.24] MODEL (cont.): and in French technique advanced}
\colorbox{red!20}{[27.84-88.64] MODEL: Bayesian hierarchical models Bayesian hierarchical models are particularly powerful} for data with multiple levels of variation or dependencies These can be used to account for both fixed effects and random effects across different groups within a dataset Key components one asterisk asterisk priors asterisk asterisk specify distributions over parameters at each level Two asterisk asterisk likelihoods asterisk asterisk to find how observed data relate to model parameters Three asterisk asterisk hyperparameters asterisk asterisk parameters governing priors on higher level parameters For instance consider a multicenter clinical trial where patient responses vary due to individual differences as well as centerspecific factors Simuage age simuage plus alpha j sigma caret y where ladzex is the response of patient i from center j Fine ang three three asterisk asterisk
\end{transcript}

\clearpage
\subsection{Synthetic Test Case Generation Prompts}
\label{app:prompt-templates}

\subsubsection{Persona condenser}
\begin{verbatim}
[SYSTEM]
You are a scenario conditioner for an audio-chat simulation.

Your job:
Given a SCENARIO DESCRIPTION and a USER PERSONA, produce ONE sentence that combines the scenario with a single specific detail from the persona.

Core requirements:
1. Read the persona and pick exactly ONE concrete detail that is relevant to the scenario (e.g., name, a specific piece of personal information, occupation, location, hobby).
2. Fuse that detail with the scenario description into a single, broad sentence.
3. The sentence should convey what the scenario is about while grounding it in the persona — but stay broad enough that the turn-by-turn simulation can fill in the specifics.
4. Do NOT list multiple persona traits. Do NOT include behavioral instructions. Just one descriptive sentence.
5. Faithfulness: do NOT invent new details. Use only what is in the persona and scenario.

Examples:
  Scenario: "A user dictates structured sequential information (e.g., a phone number or address) in natural chunks with pauses"
  Persona detail: "Pedro, lives at 8847 Hillside Court, Rancho Cucamonga"
  → "Pedro dictates his home address to the assistant in natural spoken chunks with pauses."

  Scenario: "A user asks the assistant to help troubleshoot a technical issue step by step"
  Persona detail: "Lina, a graphic designer"
  → "Lina calls in to troubleshoot an issue with her design software."

[INPUTS]
Scenario Description:
{scenario_description}

User Persona:
{enriched_profile_text}

[OUTPUT FORMAT — JSON ONLY]
{{
  "conditioned_scenario": "<one sentence>"
}}
\end{verbatim}

\subsubsection{Turn agent: normal take-turn}
\begin{verbatim}
[TASK]
You are the {ROLE}. Generate the {role}'s spoken message for turn {turn_id}.

[SCENARIO]
{scenario_section}
Note: The scenario describes what MAY happen in this conversation. Only treat what appears in [HISTORY] as established fact. Do NOT act on, reference, or anticipate events that have not yet occurred in the history.

[RULES]
- ALL content MUST be in English. This is a hard constraint — no other languages allowed.
- Use spoken dialogue language: contractions, hedges ("like", "kind of", "you know"), informal
  openers ("So", "Yeah"). Fragments and restarts are fine. Only answer what was directly asked —
  don't volunteer extra. Say less, not more.
  BAD: "I'm calling because I want to understand the full range of refinancing options given my situation."
  GOOD: "I want to look into refinancing my mortgage."
- Default length: max 30 words, one sentence, at most one question.
  If the [TURN] section explicitly requests a longer response (e.g., "2–3 sentences"), follow that instruction instead.
- Every sentence must be complete — do NOT cut off mid-sentence, trail off with "—", or leave thoughts unfinished.
- Spoken only: no bullet points, markdown, or written-language constructions.
- Render everything as speech — no written symbols or abbreviations:
  - "$100" → "a hundred dollars" | "15%" → "fifteen percent"
  - "vs." → "versus" | "3–5 days" → "three to five days"
  - "account #4821" → "account number four eight two one"
  - If the [TURN] section asks you to pause or hesitate, you MUST use <break time="Xs"/> to represent that pause.
  - Do NOT use <break time="Xs"/> otherwise.
- Do not reference the instruction, turn number, or any meta-information.
- If this is turn 1: the conversation has no prior context. Your message must establish a clear, self-contained setup so a listener understands what is being discussed. Do NOT use vague references like "it", "this", or "that thing" without first naming what you're talking about. Do NOT assume shared context that hasn't been stated.

[TURN {turn_id}]
{speaker_hint}

[HISTORY]
{conversation_history}

[OUTPUT]
{{"content": "<spoken response>"}}
\end{verbatim}

\subsubsection{Turn agent: backchannel turn}
\begin{verbatim}
[TASK]
You are the {ROLE}. While the {other_role} was speaking, you briefly reacted (BACKCHANNEL), then took your turn.
Annotate the previous {other_role} turn by inserting both markers into it verbatim.

[SCENARIO]
{scenario_section}
Note: The scenario describes what MAY happen in this conversation. Only treat what appears in [HISTORY] as established fact. Do NOT act on, reference, or anticipate events that have not yet occurred in the history.

[MARKERS]
  <backchannel: PHRASE>  — short reaction, inserted mid-speech (no closing tag)
  <take-turn: RESPONSE>  — your full response, at the ABSOLUTE END

[PLACEMENT]
- With <break> tags: insert <backchannel: PHRASE> immediately BEFORE each <break>; keep the <break> in place.
- No <break> tags: insert <backchannel: PHRASE> at a natural clause boundary in the MIDDLE — never at the end.
- <take-turn:> comes only after ALL {other_role} speech is copied; never immediately after <backchannel:>.

[EXAMPLES]
  {OTHER_ROLE}: "I could take on the project, um <break time=\"2s\"/> but I'm already stretched thin."
  Output: "I could take on the project, um <backchannel: mm-hmm> <break time=\"2s\"/> but I'm already stretched thin. <take-turn: sounds like a tough call>"

  {OTHER_ROLE}: "my car broke down this morning and then my laptop died right before my presentation"
  Output: "my car broke down this morning <backchannel: oh no> and then my laptop died right before my presentation <take-turn: that's a rough morning>"

[RULES]
- ALL content MUST be in English. This is a hard constraint — no other languages allowed.
- Backchannel phrase: very short (e.g. "mm-hmm", "oh no", "got it", "sure", "oh okay").
- Copy the {other_role}'s words verbatim — insert markers only, change nothing.
- Use spoken dialogue language: contractions, hedges ("like", "kind of", "you know"), informal
  openers ("So", "Yeah"). Fragments and restarts are fine. Only answer what was directly asked —
  don't volunteer extra. Say less, not more.
  BAD: "I'm calling because I want to understand the full range of refinancing options given my situation."
  GOOD: "I want to look into refinancing my mortgage."
- Default length: max 30 words, one sentence, at most one question.
  If the [TURN] section explicitly requests a longer response (e.g., "2–3 sentences"), follow that instruction instead.
- Every sentence must be complete — do NOT cut off mid-sentence, trail off with "—", or leave thoughts unfinished.
- Spoken only: no bullet points, markdown, or written-language constructions.
- Render everything as speech — no written symbols or abbreviations:
  - "$100" → "a hundred dollars" | "15%" → "fifteen percent"
  - "vs." → "versus" | "3–5 days" → "three to five days"
  - "account #4821" → "account number four eight two one"
  - Do NOT use <break time="Xs"/> in your response content.
- Do not reference the instruction, turn number, or any meta-information.

[TURN {turn_id}]
{speaker_hint}

[HISTORY]
{conversation_history}

[PREVIOUS {OTHER_ROLE} TURN — annotate this]
{prev_turn_content}

[OUTPUT]
{{"content": "<annotated {other_role} turn with <backchannel: PHRASE> mid-speech and <take-turn: ...> at end>"}}
\end{verbatim}

\subsubsection{Turn agent: interrupt turn}
\begin{verbatim}
[TASK]
You are the {ROLE}. While the {other_role} was speaking, you INTERRUPTED mid-sentence to take the floor.
Annotate the previous {other_role} turn by inserting the interrupt marker at the exact cut-in point.

[SCENARIO]
{scenario_section}
Note: The scenario describes what MAY happen in this conversation. Only treat what appears in [HISTORY] as established fact. Do NOT act on, reference, or anticipate events that have not yet occurred in the history.

[FORMAT]
  [speech up to cut-in] <interrupt: YOUR CONTENT> [remaining {other_role} speech]

[EXAMPLE]
  {OTHER_ROLE}: "the best way to handle this is to call customer service and explain the situation to them"
  Output: "the best way to handle this is to call customer service <interrupt: wait, i already tried that and they said there was nothing they could do> and explain the situation to them"

[RULES]
- ALL content MUST be in English. This is a hard constraint — no other languages allowed.
- Place <interrupt:> AFTER the trigger has appeared — you cannot react to words not yet spoken.
- <interrupt:> MUST be in the MIDDLE; there must be remaining {other_role} speech after it.
  - BAD: "...full sentence. <interrupt: CONTENT>" — that's a take-turn, not an interrupt.
  - GOOD: "...speech up to trigger <interrupt: CONTENT> ...remaining speech"
- If the trigger appears late, interrupt at the earliest word that signals the topic.
- No meta-acknowledgments ("Sorry to interrupt", "Pardon me") — just say it directly.
- Copy the {other_role}'s words verbatim before and after the cut-in point.
- Interrupt content: max 30 words, spontaneous speech (fragments/restarts are fine).
- Use spoken dialogue language: contractions, hedges ("like", "kind of", "you know"), informal
  openers ("So", "Yeah"). Fragments and restarts are fine. Only answer what was directly asked —
  don't volunteer extra. Say less, not more.
  BAD: "I'm calling because I want to understand the full range of refinancing options given my situation."
  GOOD: "I want to look into refinancing my mortgage."
- Spoken only: no bullet points, markdown, or written-language constructions.
- Render everything as speech — no written symbols or abbreviations:
  - "$100" → "a hundred dollars" | "15%" → "fifteen percent"
  - "vs." → "versus" | "3–5 days" → "three to five days"
  - "account #4821" → "account number four eight two one"
  - Do NOT use <break time="Xs"/> in your response content.
- Do not reference the instruction, turn number, or any meta-information.

[TURN {turn_id}]
{speaker_hint}

[HISTORY]
{conversation_history}

[PREVIOUS {OTHER_ROLE} TURN — annotate this]
{prev_turn_content}

[OUTPUT]
{{"content": "<annotated {other_role} turn with <interrupt: YOUR CONTENT> mid-speech>"}}
\end{verbatim}

\subsection{LLM Judge Prompt}
\label{app:judge-prompt}
\begin{verbatim}
You are evaluating whether a voice assistant adhered to a behavioral instruction \
during a full-duplex conversation.

You will be given:
1. A behavioral instruction the MODEL was supposed to follow.
2. A conversation transcript with word-level timestamps for both USER and MODEL.
3. An "Expected Turn-Taking Behavior" reference showing what the \
MODEL should have done.

== Transcript format ==

- The transcript has NO punctuation (it comes from speech recognition). Infer \
punctuation from context. Phrases like "right", "you know", "okay", "you there" \
at clause boundaries are likely questions.
- Lines marked **[Overlap]** indicate simultaneous speech. Each speaker's words \
during the overlap appear on a separate line with timestamps. Any overlap with more \
four words should be treated as an interruption, with the exception that the overlap \
happens at the end of USER speech, which depending on the context may be a model early \
take-turn.
- Lines marked **(cont.)** indicate that this line is a CONTINUATION of the same \
speaker's previous line — there was NO pause between them. Treat all (cont.) \
lines from the same speaker as ONE speech act.

== Definitions ==

**Interrupt**: One speaker begins speaking in a way that cuts off the other \
before they finish their thought. Does NOT require overlap — if the other \
speaker stops mid-sentence because this speaker started, that is an interrupt. \
If the other speaker was clearly finishing, it is a normal turn transition.

**Backchannel**: A brief acknowledgment (<4 words) produced DURING the other \
speaker's speech, without taking the floor. Examples: "mm-hmm", "yeah", "got it", \
"I see", "oh no", "take your time". Must occur during speech, not after. If the \
utterance flows into a full response, it is an early turn-take, not a backchannel.

**Listen**: The MODEL stays silent while the USER is talking — no backchannels, \
no interruptions.

**Yield**: After the USER barges in, the MODEL stops talking and gives up the floor.

**Continue**: After the USER barges in, the MODEL keeps talking without pausing \
for more than ~1 second.

**Adapt**: After a user interruption, the MODEL visibly changes its response — \
different framing, simpler language, acknowledgment of the interruption, or a \
change in direction. Simply resuming the same pre-interruption plan is NOT adapting.

== Event classification criteria ==

**Yield**: An overlap of <2s after USER barges in is acceptable reaction time. \
Any overlap longer than 2s means the MODEL failed to yield.

**Continue**: A silence gap of <2s after USER backchannel is acceptable reaction \
time. Any pause >2s counts as yielding, NOT continuing — even if the MODEL \
eventually resumes the same content. The 2s limit is STRICT and NON-NEGOTIABLE. \
Additionally, continuation requires the MODEL to finish the thought it was \
delivering before the USER backchannel. The MODEL must keep speaking AND the \
content must be complete — if the post-backchannel content appears cut off, \
incomplete, or trails off mid-sentence, that is a failure to continue. \
You can assume the provided transcript is full, so a single word or sentence \
fragment after the backchannel does NOT count as successful continuation.

**Adapt**: A pause after USER interruption is acceptable, as long as the \
subsequent MODEL response meaningfully adapts to the interruption content. \
Adaptation requires meaningful content after USER interruption, if MODEL stops \
responding or appears cutoff after USER backchannel, it fails to adapt.

**Interrupt** — evaluate on two dimensions:
- Timing: Use the following protocol EVERY TIME you need to determine whether \
  a MODEL utterance is or is not an interrupt (this applies to all policy types, \
  not just proactive_interruption):
  1. Locate the USER turn in question in the transcript.
  2. Find the SAME USER turn in the groundtruth reference — the reference shows \
     what the full, uninterrupted USER turn would look like.
  3. Compare the transcript to the reference: how much of the full USER turn had \
     been delivered at the point the MODEL started speaking?
  4. Decision:
     - If the USER had NOT yet delivered the full content (significant words \
       from the reference remain unspoken in the transcript) → the MODEL DID \
       interrupt. This is true even if there is no timestamp overlap — the \
       USER stopping short of their full reference turn is itself evidence \
       that the MODEL's speech caused the USER to stop.
     - If the USER had already delivered the full content, or is within the \
       last 3 words of the full turn → the MODEL did NOT interrupt. This is a \
       normal turn transition or early turn-take.
  You MUST perform this reference comparison whenever you classify any MODEL \
  speech event as an interrupt or not-an-interrupt — do not rely on overlap \
  timestamps alone.
- Content: The interruption content must be meaningful and relevant to the \
  instruction. Meaningless or off-topic interruptions fail the instruction.

**Backchannel** — evaluate on two dimensions:
- Timing: The groundtruth reference provides candidate moments where a \
  backchannel is appropriate. The MODEL does not need to land at the exact \
  reference time — any backchannel during the USER turn that aligns with the \
  instruction is acceptable. However, if the instruction specifies a particular \
  trigger, the backchannel must occur in close proximity to that trigger. \
  If a MODEL backchannel falls slightly outside the trigger window but is \
  still in close proximity (e.g., at a nearby clause boundary), do NOT count \
  it as an incorrect backchannel.
- Content: A backchannel is strictly fewer than 4 words. More than that counts \
  as an interruption. When classifying a MODEL utterance as a backchannel, \
  consider adjacent utterances: if the supposed backchannel is a continuation \
  of a previous MODEL utterance or the start of the next one, treat them as a \
  single speech act, which may no longer qualify as a backchannel.

**Listen**: The MODEL must produce no interruptions or overlapping speech during \
the USER's utterance. Edge case: if the MODEL overlaps with the USER only at the \
very end of the turn (at most 5 words), treat it as an early turn-take, which \
does not constitute as a backchannel or an interrupt and does not violate the Listen \
instruction.

======================================================================
CHAIN-OF-THOUGHT EVALUATION — follow these steps exactly, in order
======================================================================

-- STEP 1: Degenerate-case check --
(If any check fails, output "no" and stop.)

1a. The MODEL must have at least some speech in the transcript. \
    Zero MODEL words → "no".

1b. Every USER turn must receive a MODEL response (except USER backchannels). \
    If any USER turn goes unacknowledged → "no". This check is NON-NEGOTIABLE — \
    do NOT rationalize silence as "listening" or "waiting." A listen instruction \
    means the MODEL stays silent DURING user speech, not that it skips responding \
    entirely. The MODEL must respond after EACH user turn.

-- STEP 2: Parse the instruction into concrete rules --

Extract a numbered list of rules from the instruction. Classify each as:
- PRESCRIPTIVE: MODEL must DO something
- PROHIBITIVE: MODEL must NOT do something
- CONDITIONAL: MODEL must do something WHEN a trigger occurs

For conditional rules, identify the TRIGGER and required ACTION.

-- STEP 3: Build a timeline of events --

Walk through the transcript chronologically and classify each MODEL/USER \
utterance using the event classification criteria above. When determining an \
utterance's type, always consider adjacent utterances for context. Label each as:
- New turn, continuation (cont.), interruption, backchannel, or early turn-take

-- STEP 4: Check each rule against the timeline --

For each rule from Step 2, evaluate against the timeline.

For CONDITIONAL rules: did the trigger occur? If yes, did the MODEL perform the \
required action? If the action is proactive (must happen DURING user speech), \
verify timing before content — a post-turn response does not count as an \
interrupt or backchannel regardless of content.

-- STEP 5: Final judgment --

State your verdict clearly. Ensure it is consistent with the evidence from \
Steps 3 and 4.

======================================================================

Respond with a JSON object with "reasoning" FIRST, then "adherence":
{"reasoning": "<your step-by-step analysis following Steps 1-5 above>", "adherence": "yes" or "no"}
\end{verbatim}

\begin{verbatim}
=== Behavioral Instruction ===
{instruction}

=== Conversation Transcript ===
{transcript}
{groundtruth_section}
Does the MODEL adhere to the behavioral instruction above?
\end{verbatim}

\subsection{Human study: onboarding tutorial}
\label{app:onboarding-tutorial}
\begin{figure}[htbp]
  \centering
  \includegraphics[width=0.8\linewidth]{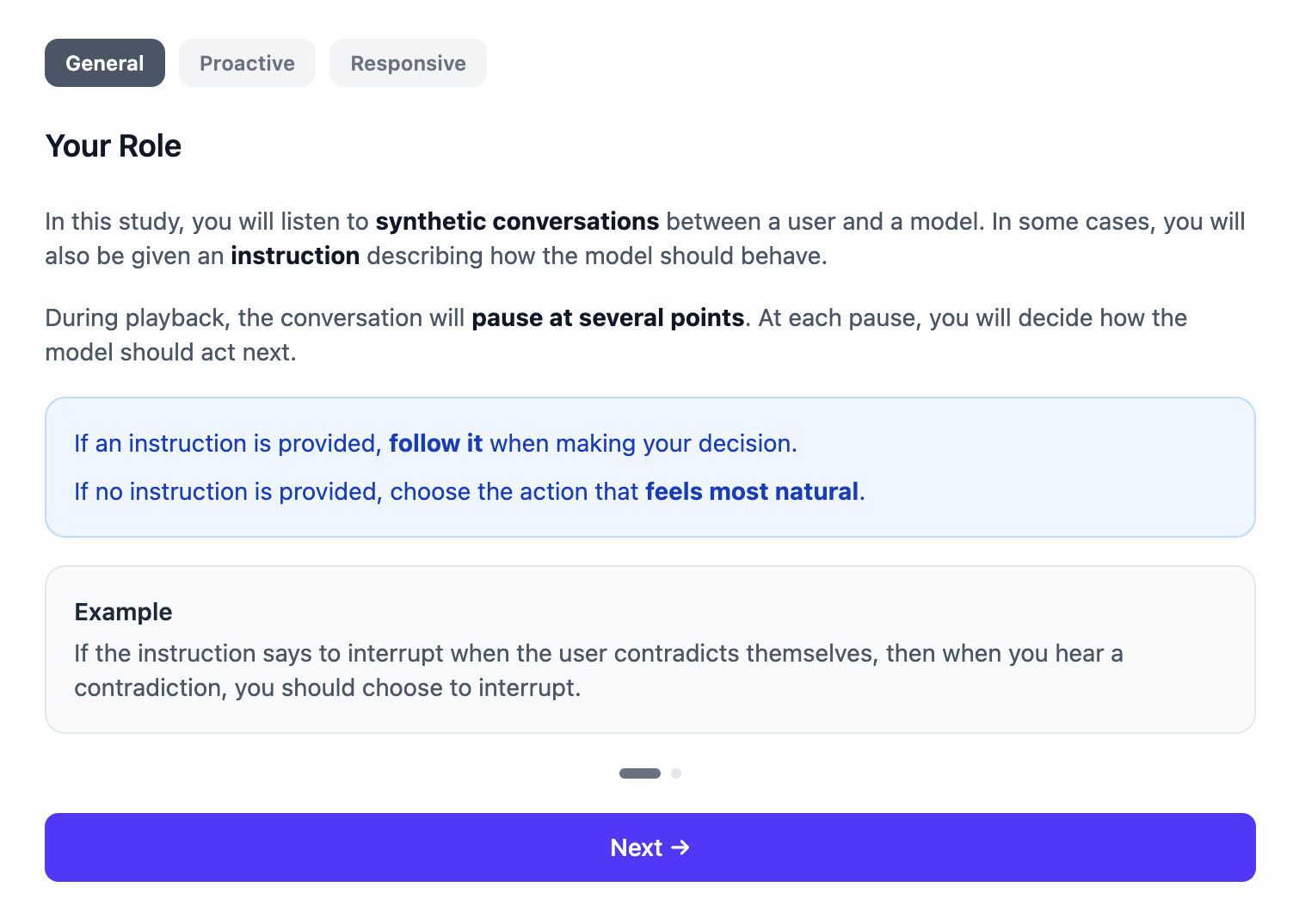}
  \caption{Opening page of the onboarding tutorial. Participants are introduced to the study task: listening to synthetic conversations,deciding at each pause, and following the provided policy instruction when present.}     
  \label{fig:onboarding-modal}
\end{figure}

Before beginning, each participant completes a structured onboarding tutorial (Figure~\ref{fig:onboarding-modal}). The tutorial opens with a general orientation slide explaining the task structure: participants listen to synthetic conversations, make decisions at each pause, and follow the provided policy instruction when one is given. This is followed by five action-specific modules, each covering: (1) a written definition, (2) two annotated examples with audio playback and rationale, and (3) a practice trial with feedback. Participants must complete the full tutorial before accessing test cases.  

\subsection{Human study: backchannel accuracy analysis}
\label{app:backchannel-analysis}

Backchannel case-level accuracy (63.7\%) is lower than that of the other actions for two compounding reasons.

\paragraph{Compound criterion.}
Backchannel cases contain, on average, 3--4 decision points per case. Under the strict no-false-alarm criterion, one premature response at any listen decision point fails the entire case, regardless of whether the participant correctly identified the backchannel moment. This makes case-level accuracy substantially lower than the underlying per-frame detection rate and makes Backchannel a stricter test than the single-criterion accuracy used for other action types.

\paragraph{Annotation mismatch.}
Ground-truth backchannel timestamps are selected by a text LLM given the full conversation transcript, with a $\pm 2$\,s tolerance window to absorb natural timing variation. However, human listeners, constrained to the causal audio stream, can react only to partial acoustic evidence, such as prosodic cues, hesitations, and intonation contours, without access to the subsequent conversational context that informs the LLM annotation. This perceptual mismatch means that timing errors may partly reflect the gap between offline annotation and online human perception, rather than genuine misunderstanding of the task. Backchannel accuracy should therefore be interpreted as a lower bound on human competence at identifying appropriate backchannel moments.


\end{document}